\documentclass{ecai}

\usepackage{latexsym}
\usepackage{amssymb}
\usepackage{amsmath}
\usepackage{amsthm}
\usepackage{booktabs}
\usepackage{enumitem}
\usepackage{graphicx}
\usepackage{color}
\usepackage{microtype}
\usepackage{hyperref}
\usepackage[htt]{hyphenat}

\usepackage{amsmath}
\usepackage{amsthm}
\usepackage{amssymb}
\usepackage{amsfonts}

\usepackage{booktabs}
\usepackage{tabularx}
\usepackage{tabularray}
\usepackage{longtable}

\usepackage{placeins}
\usepackage{enumitem}

\usepackage{wrapfig} 

\usepackage{algorithm}
\usepackage{algorithmic}

\usepackage[switch]{lineno}

\newcommand{\BibTeX}{B\kern-.05em{\sc i\kern-.025em b}\kern-.08em\TeX}

\usepackage{cleveref}
\crefname{figure}{figure}{figures}

\hypersetup{
		pdfencoding=auto, 
		psdextra,
		colorlinks=true,
		citecolor=green!40!black,
		linkcolor=red!50!black,
		urlcolor=blue!80!black
	}
\usepackage{subcaption}

\let\oldemph\emph
\renewcommand{\emph}[1]{\textcolor{blue!60!black}{\oldemph{#1}}}

\usepackage{fancyhdr}
\usepackage{mathtools}
\usepackage{pifont}
\usepackage{subcaption}
\usepackage{booktabs}
\usepackage{multirow}
\usepackage[export]{adjustbox}
\usepackage{xargs}
\usepackage{xcolor}

\usepackage{booktabs}
\usepackage{nicefrac}
\usepackage{chngpage}
\usepackage{todonotes}
\usepackage{cleveref}
\usepackage{placeins}
\usepackage{wrapfig}
\usepackage{titlesec}

\titlespacing*{\section}
{0pt}{2.4ex plus 0.8ex minus .2ex}{1.2ex plus .2ex}

\titlespacing*{\subsection}
{0pt}{2.2ex plus 0.6ex minus .2ex}{0.8ex plus .2ex}

\titlespacing*{\subsubsection}
{0pt}{1.2ex plus 0.6ex minus .2ex}{0.6ex plus .2ex}

\theoremstyle{remark}

\theoremstyle{plain}

\numberwithin{equation}{section}

\usepackage{todonotes}

\makeatletter 
\def\WFfill{\par 
    \ifx\parshape\WF@fudgeparshape 
    \nobreak 
    \ifnum\c@WF@wrappedlines>\@ne 
    \advance\c@WF@wrappedlines\m@ne 
    \vskip\c@WF@wrappedlines\baselineskip 
    \global\c@WF@wrappedlines\z@ 
    \fi 
    \allowbreak 
    \WF@finale 
    \fi 
} 
\makeatother

\newcolumntype{R}{>{\raggedleft\arraybackslash}X}
\newcolumntype{Z}{>{\centering\arraybackslash}X}
\newcolumntype{W}{>{\centering\arraybackslash}m{.75in}}

\definecolor{darkgreen}{RGB}{1,50,32}

\begin{document}

\begin{frontmatter}

\title{Is Textual Similarity Invariant under Machine Translation? Evidence Based on the Political Manifesto Corpus}

\author[A]{Daria Boratyn}
\author[A]{Damian Brzyski}
\author[A]{Albert Leśniak}
\author[A]{Wojciech Łukasik}
\author[B]{Maciej Rapacz}
\author[C]{Jan Rybicki}
\author[A]{Wojciech Słomczyński}
\author[A]{Dariusz Stolicki \thanks{Corresponding Author. Email: dariusz.stolicki@uj.edu.pl.\\ This research has been funded under the Polish National Center for Science grant no.~2023/49/B/HS5/03893 and the Jagiellonian University Excellence Initiative, DigiWorld PRA, QuantPol Center project.}}

\address[A]{Jagiellonian Center for Quantitative Political Science, Jagiellonian University, Kraków, Poland}
\address[B]{AGH University, Kraków, Poland}
\address[C]{Jagiellonian Center for Digital Humanities, Jagiellonian University, Kraków, Poland}

\begin{abstract}
We investigate the extent to which cosine similarity between paragraph
embeddings is invariant under machine translation, using the Manifesto
Corpus of over 2,800 political party platforms in 28 languages translated
to English via the EU eTranslation service. Rather than measuring
translation-induced semantic shift directly we measure the stability
of pairwise similarity relationships across embedding models, and use
inter-model disagreement on original-language text as a calibrated
invariance threshold. This yields a per-language non-inferiority test
for four hypotheses about how translation interacts with embedding choice,
with verdicts that distinguish languages where translation demonstrably
preserves semantic structure from those where it demonstrably degrades it
and from those where the available evidence does not resolve the question.
The framework is corpus- and pipeline-agnostic and extends naturally to
downstream tasks. Applied to our data, it identifies ten languages with
translation invariance and four with detectable distortion.
\end{abstract}

\end{frontmatter}

\section{Preliminaries}

Measuring semantic similarity between texts is a fundamental NLP problem underlying many applications, from information retrieval, semantic search, and document organization, through recommendation systems and paraphrase detection, to digital humanities and computational social science applications like ideological positioning, authorship attribution, and genre recognition. While numerous methods exist for same-language comparison, cross-lingual similarity poses unique challenges. One common solution is to translate all texts into a single language before applying standard comparison methods. For applications involving large numbers of texts, translation necessarily means `machine translation.'

But do similarity relationships between texts remain preserved when translating them into another language, and can modern machine translation models maintain these relationships? In the present paper, we investigate these questions using the EU eTranslation model and various common approaches to similarity measurement.

\subsection{Motivation}

Cross-lingual applications—ranging from plagiarism detection and information retrieval to large-scale digital humanities—are predicated on the assumption that we can measure ``how similar'' two pieces of text are even when they are written in different languages. Unfortunately, the tools that work well in a purely monolingual setting do not transfer automatically. Monolingual sentence- or document-level embeddings live in language-specific vector spaces whose axes are not aligned; Euclidean or cosine distances computed across those spaces therefore lack any linguistic meaning.

Current practice tackles this misalignment in two broad ways. The first strategy is pivot translation: every text is machine-translated into a single hub language and similarity is computed there. The attraction is obvious—once in the same language, we can deploy the full power of state-of-the-art monolingual models—but translation noise can warp the very semantic relations we hope to study, especially for low-resource or stylistically marked inputs \citep{Rybicki23}. The second strategy is to build intrinsically multilingual representations, typically by pre-training a Transformer on parallel or comparable corpora. These models offer language-agnostic embeddings out of the box, but they still trail their monolingual counterparts in fine-grained semantic tasks and may encode subtle language-specific biases \citep{BlevinsEtAl24}.

In this paper we seek to determine whether the trade-off is as stark as it appears. We find that, for a broad family of semantic similarity measures, the relative geometry of texts is somewhat stable under high-quality machine translation -- but not as stable as for multilingual models. By quantifying this stability across languages, we provide concrete evidence for determining when translation-based pipelines are sufficient, and when more computationally expensive multilingual models are actually needed.

\subsection{Contribution}

We evaluate the effectiveness of machine translation for cross-lingual text similarity analysis by comparing similarity matrices derived from original texts against those from their machine-translated versions. Our contribution is twofold: methodologically, we introduce a novel measure of similarity invariance, as well as several auxiliary measures based on downstream task performance; empirically, we apply these measures to a comprehensive corpus of political party programs. We benchmark the machine translation approach against several state-of-the-art multilingual large language models (LLMs) to provide practical guidance for cross-lingual text analysis.

\subsection{Applications}

Our results are of the most immediate interest to researchers seeking to analyze multilingual text corpora, whom we can provide with a recommendation as to which approach to measuring similarity in such corpora best preserves original-language similarity structure, as well as with an assessment of how much distortion they should expect.

Our findings are of immediate, practical interest primarily for researchers who work with mixed-language data, for instance digital humanities scholars mapping thematic networks across national literature corpora, computational social scientists tracking how news narratives migrate between language communities, or product teams that need a single relevance-ranking pipeline for a multilingual search index. Because we quantify how closely semantic similarity matrices survive translation, researchers can now choose a pipeline by matching our reported invariance scores to their own tolerance for distortion: when similarity preservation is sufficiently high, a translation-plus-monolingual-encoder workflow is likely to be both cheaper and more accurate than a multilingual model fine-tuned from scratch.

\subsection{State-of-the-Art}

Modern neural machine translation (NMT) systems -- especially Transformer-based models since 2017 -- have dramatically advanced translation quality. However, research from 2018 to 2025 reveals that machine-generated translations still exhibit distinct linguistic footprint known as \emph{translationese} \citep{Gellerstam86,KurokawaEtAl09,RileyEtAl20}. These traits include awkward or unidiomatic phrasing due to overly literal renderings of the source, as well as systematic differences in lexical choice and syntax \citep{VolanskyEtAl15}. Usually, translationese effects are stylistic rather than semantic. For instance, translated documents carry over source-language influences, altering the distribution of words in the target language \citep{ChowdhuryEtAl20,Rybicki23}. Translations generally use a narrower vocabulary and more repetitive word choice compared to original text, with NMT outputs being even more lexically limited than human translations \citep{Toral19,NiuJiang24}. They tend to introduce greater explicitation, adding clarifying words or connectors that were implicit in the source \citep{Baker93,RileyEtAl20}. Finally, translations frequently use shorter sentences or simpler grammatical structures than the source \citep{Baker93}, although this particular effect appears for NMT less frequently than for human translators \citep{NiuJiang24,FuLiu24}.

Of greater concern to researchers, especially those using machine translation in their text processing pipelines, are \emph{semantic shifts}, i.e., changes resulting in an alteration or loss of meaning. Those include not only mistranslation, but also under-translation, i.e., omission of words or phrases; over-translation, i.e., addition of content not present in the source; syntactic modifications that affect meaning; and logical inconsistencies \citep{HeEtAl21}. More subtle shifts include failure to convey the sentiment polarity of the source \citep{SaadanyOrasan20}. Such shifts are disproportionately frequent for low-resource languages \citep{AgrawalEtAl24}. Indeed, NMT research reports that for languages with scant training data, critical translation errors (omissions, mistranslations) are far more prevalent, often requiring extensive post-editing \cite{SaadanyOrasan20}. In such cases, vital information (like negation or modality) may be mistranslated, directly affecting the semantic correctness of the output.

To quantify semantic preservation, authors use a variety of metrics. For instance, \citet{KhobragadeEtAl19} proposed an entailment-based MT evaluation metric: a perfect translation should be mutually entailing with the source. Other works use round-trip translation as a diagnostic: translate from source to target and back to see if the original meaning is recovered. Discrepancies can reveal shifts in meaning. The referential transparency method by \citet{HeEtAl21} created pairs of sentences that should translate the same (differing only in context) and used them to catch inconsistent translations. The ongoing research in MT evaluation (e.g., the ACES challenge sets \citep{AmrheinEtAl22}) is increasingly focused on these fine-grained semantic fidelity issues, moving beyond coarse metrics to identify specific weaknesses in conveying meaning.

The methods surveyed above all share a common methodological commitment: they attempt to identify or quantify translation-induced semantic shift directly, by comparing source and target texts on intrinsic measures of meaning preservation -- whether through entailment-based scoring \citep{KhobragadeEtAl19}, round-trip translation diagnostics \citep{MoonEtAl20}, referential transparency testing \citep{HeEtAl21}, or fine-grained challenge sets that probe specific failure modes \citep{AmrheinEtAl22}. These approaches require, implicitly or explicitly, an external standard of semantic equivalence against which translated text can be judged: a gold-standard reference translation, a logically-entailing rephrasing, or a set of pre-curated contrasts whose preservation under translation is itself the test. Such standards are difficult to construct at scale and across many languages, and the resulting evaluations are themselves bounded by the quality and coverage of the reference material.

We pursue an alternative strategy that requires no external semantic gold standard: rather than asking whether the meaning of any particular text is preserved under translation, we ask whether the \emph{structure of similarity relationships} among many texts is preserved. This shifts the inferential target from absolute semantic equivalence between an original and a translation -- which no embedding model perfectly captures and no automated metric perfectly detects -- to the relative geometry of an entire corpus of texts in embedding space, which is directly measurable from the embeddings themselves and admits calibration against the natural variability of embedding models on the original-language texts.

\section{Data and Methods}

To test our hypothesis of the translation-invariance of semantic similarity, we proceed in six stages:
\begin{enumerate}
\item We start with a multilingual corpus of texts from a single domain (political party programs -- see Subsection \ref{sec:data} for a more detailed discussion).
\item Every non-English text in the corpus is machine-translated into English (see Subsection \ref{sec:xlat}). This step creates a parallel English corpus that will be compared to the original-language corpus.
\item Every text (both in the original corpus and the translated corpus) is split into sentences and pseudo-paragraphs.
\item Paragraph embeddings are obtained by computing sentence embeddings and pooling them for each pseudo-paragraph. We use language-specific models for the original corpus, English language models for the translated corpus, and multilingual models for both corpora as a reference.
\item Within each single-language corpus we calculate the similarity matrix using cosine similarity of paragraph embeddings. We obtain one similarity matrix for each model used to calculate the embeddings.
\item We calculate correlation coefficients between per-model similarity matrices, and compare the distribution of those coefficients for models run on translated texts with the distribution for models run on original texts.
\end{enumerate}

\subsection{Data}\label{sec:data}

Our reference data set is the Party Manifesto Corpus \citep{MerzEtal16,LehmannEtAl23}, which contains political party programs from $60$ countries, totaling $3166$ texts in $39$ languages. For our analysis, we use a subset of $2878$ texts that are written in the $28$ languages supported by the EU eTranslation service -- an average of $102.79$ per language. However, the text count distribution is highly skewed, and for a majority of languages we have fewer than $49$ texts (see Table \ref{tbl:descStats} for details).

\begin{table}[h]
\begin{tabularx}{\columnwidth}{Xrrr}
\toprule
\multicolumn{1}{c}{\textbf{language}} & \multicolumn{1}{c}{\textbf{\# documents}} & \multicolumn{1}{c}{\textbf{\# sentences}} & \multicolumn{1}{c}{\textbf{\# characters}} \\ \midrule
Bulgarian & 21 & 10,125 & 2,533,239 \\
Croatian & 54 & 23,427 & 4,072,735 \\
Czech & 31 & 19,804 & 2,702,035 \\
Danish & 194 & 40,705 & 3,987,575 \\
Dutch & 231 & 343,049 & 36,520,246 \\
English & 457 & 309,776 & 41,083,372 \\
Estonian & 24 & 13,657 & 1,752,218 \\
Finnish & 98 & 26,353 & 2,948,924 \\
French & 175 & 191,391 & 28,328,047 \\
German & 269 & 208,005 & 27,256,439 \\
Greek & 69 & 42,342 & 11,353,271 \\
Hungarian & 28 & 32,041 & 5,193,132 \\
Icelandic & 42 & 7,409 & 996,961 \\
Italian & 112 & 49,624 & 9,658,649 \\
Japanese & 9 & 2,354 & 471,045 \\
Latvian & 30 & 1,071 & 140,165 \\
Lithuanian & 42 & 36,108 & 5,808,382 \\
Norwegian & 126 & 171,801 & 19,101,671 \\
Polish & 37 & 25,216 & 3,879,098 \\
Portuguese & 102 & 101,718 & 18,197,733 \\
Romanian & 39 & 9,526 & 1,979,379 \\
Russian & 36 & 7,584 & 1,953,091 \\
Slovak & 44 & 29,055 & 4,394,382 \\
Slovenian & 41 & 26,945 & 4,900,469 \\
Spanish & 356 & 379,747 & 70,346,378 \\
Swedish & 127 & 28,027 & 2,790,231 \\
Turkish & 28 & 43,974 & 6,874,307 \\
Ukrainian & 56 & 4,353 & 1,098,708 \\ \bottomrule
\end{tabularx}
\caption{Party Manifesto Corpus -- descriptive statistics.}
\label{tbl:descStats}
\end{table}

\subsection{Translation} \label{sec:xlat}

Each non-English text is machine-translated into English using the European Commission's eTranslation service \citep{OraveczEtAl22}. We use eTranslation rather than commercial alternatives as it offers unrestricted access to academic researchers without usage quotas or fees. Moreover, it offers particularly good coverage of the Manifesto Corpus languages.

\subsection{Embeddings} \label{sec:embed}

Measuring textual similarity requires transforming text into vector representations. State-of-the-art approach here is to use text embeddings, i.e., functions mapping text fragments (words, sentences, etc.) to points in some high-dimensional Hilbert space where semantic proximity corresponds to spatial proximity and arithmetic operations -- vector addition and multiplication by scalar -- have a natural interpretation in terms of semantic relationships \citep{MikolovEtAl13a}.

Modern word embedding methods are derived from the dynamically computed, contextualized hidden states of a neural network-based large language model. Such models are almost invariably based on the self-attention-based network architecture known as \emph{transformers} \citep{VaswaniEtAl17}. The two original LLM architectures were \emph{BERT} \citep{DevlinEtAl18} and \emph{GPT} \citep{RadfordEtAl18}. Current state-of-the-art architectures operate at the sentence level by pooling component word embeddings. They include \emph{SentenceBERT} \citep{ReimersGurevych19}, \emph{Universal Sentence Encoder} \citep{CerEtAl18}, \emph{AnglE} \citep{LiLi24}, \emph{LaBSE} \citep{FengEtAl22}, \emph{Mistral} \citep{JiangEtAl23}, and \emph{LLaMA} \citep{TouvronEtAl23}.

For each language (other than English) represented in our document set, we have chosen four to six models from the HuggingFace repository, usually based on \emph{SentenceBERT} architecture. For some languages, sufficient number of sentence embedding models was not available, in which case we used contextual word embedding models with mean pooling within each sentence. For model selection we relied on HuggingFace model popularity data, as well as on publicly available benchmarks, such as language-specific MTEB leaderboards \cite{MuennighoffEtAl23}. For a complete list of models used, see Appendix 1.

For translated texts, we use six state-of-the-art embedding models:
\begin{itemize}
    \item \textbf{SentenceBERT} (\texttt{sentence-transformers/stsb-bert-large}), by \citet{ReimersGurevych19}, based on BERT-large (340 M), trained on English Semantic Textual Similarity Benchmark \citep{CerEtAl17}, $1024$-dimensional output, STSB score: $0.8445$;
    \item \textbf{SMPNet} (\texttt{sentence-transformers/nli-mpnet-base-v2}) \citep{ReimersGurevych21}, developed by UKPLab, based on NLI MPNet-base (110 M), trained on 1 B+ English NLI \& STS pairs, $768$-dimensional output, MTEB English score: $63$;
    \item \textbf{Universal AnglE Embedding} (\texttt{WhereIsAI/UAE-Large-V1}), developed by \citet{LiLi24}, based on the Angle-optimized Embeddings architecture implemented on top of a RoBERTa-large (335 M), trained on, among others, Multi-Genre NLI (MNLI) \citep{WilliamsEtAl18} and Stanford NLI (SNLI) \citep{BowmanEtAl15} datasets, $1024$-dimensional output, MTEB English score: $64.64$;
    \item \textbf{E5-Mistral 7B} (\texttt{intfloat/e5-mistral-7b-instruct}), developed by \citet{WangEtAl24}, based on a 32-layer Mistral 7B large language model \citep{JiangEtAl23} fine-tuned using Microsoft's contrastive, prefix-aware E5 procedure \citep{WangEtAl22} on a mixture of synthetic and MS-MARCO data, $4096$-dimensional output, MTEB English score $67.97$. For embedding extraction, each text is prefixed with an instruction to `\texttt{retrieve semantically similar text}' and followed by an explicit EOS token. The sentence embedding is the last hidden state at the EOS-token position, computed with attention-mask-aware last-token pooling so that padding does not affect the extracted vector \citep{WangEtAl24};
    \item \textbf{LLaMA 2 7B} (\texttt{NousResearch/Llama-2-7b-hf}), developed by Meta \citep{TouvronEtAl23,TouvronEtAl23a}, based on a 32-layer 6.7B-parameter custom large language model, $4096$-dimensional output. Since LLaMA has no canonical sentence-embedding extraction procedure, we used the prompt-induced last-token method of \citet{JiangEtAl24a}: each text is wrapped in the template \texttt{Summarize sentence "\{t\}" in one word:"}, the model is run in inference mode, and the sentence embedding is taken as the last hidden state of the prompt-final token. This technique elicits a sentence-level summary representation at the position immediately preceding the prospective one-word answer;
    \item \textbf{NV-Embed 2} (\texttt{nvidia/NV-Embed-v2}), developed by Nvidia \citep{LeeEtAl24}, based on a latent attention layer, decoder-only 7.85-B parameters, $4096$-dimensional output, MTEB English score $69.81$.
\end{itemize}

In addition, we use seven state-of-the-art multilingual embedding models:

\begin{itemize}
    \item \textbf{SMPNet Multilingual} (\texttt{sentence-transformers/ paraphrase-multilingual-mpnet-base-v2}): developed by \citet{ReimersGurevych20}, based on the same NLI MPNet-base (110 M) model as SMPNet; $768$-dimensional output; MTEB Multilingual score: $51.98$;
    \item \textbf{XLM-RoBERTa} (\texttt{FacebookAI/xlm-roberta-large}), developed by Facebook AI \citep{ConneauEtAl20a}, based on RoBERTa (550 M), pre-trained on CommonCrawl; $1024$-dimensional output;
    \item \textbf{XLM-R Longformer} (\texttt{markussagen/xlm-roberta-longformer-base-4096}), developed by Markus Sagen, trained from the XLM-RoBERTa checkpoint using the Longformer pre-training scheme \citep{BeltagyEtAl20} on the English WikiText-103 corpus; $1024$-dimensional output;
    \item \textbf{LaBSE} (\texttt{sentence-transformers/LaBSE}), where LaBSE stands for Language-agnostic BERT Sentence Embedding model, developed by Google \citep{FengEtAl22}, pre-trained on Wikipedia and CommonCrawl; $768$-dimensional output; MTEB Multilingual score: $52.07$; 
    \item \textbf{mT5} (\texttt{google/mt5-large}), developed by Google \citep{XueEtAl21}, based on the Text-to-Text Transfer Transformer architecture and pre-trained on the mC4 dataset; $1024$-dimensional output;
    \item \textbf{Nomic Embed 1.5} (\texttt{nomic-ai/nomic-embed-text-v1.5}), developed by \citet{NussbaumEtAl25}, based on long (2048 tokens) context length BERT; $768$-dimensional output; MTEB Multilingual STS score: $59.45$; 
    \item \textbf{BGE-M3} (\texttt{BAAI/bge-m3}), developed by \citet{ChenEtAl24}; $1024$-dimensional output; MTEB Multilingual STS score: $74.12$.
\end{itemize}

Each of those models is run both on original-language texts and on translated ones.

\subsection{Segmentation} \label{sec:segment}

Longer texts, such as political manifestos, typically cover multiple, diverse topics. When such texts are compared at the document-level, embeddings or other representations for distinct semantic topics become conflated, resulting in regression to the mean. This, in turn, generates artificially high level of similarity across texts. To avoid this problem, we segment our texts into approximately single-topic chunks (\emph{pseudo-paragraphs}) and measure semantic similarity between those chunks, rather than whole documents.

Unfortunately, we cannot rely on original text segmentation. First, it has been removed from many texts included in the Manifesto Corpus. More importantly, political manifestos vary in formatting -- some have long paragraphs covering multiple points, while others use short bullet points. To ensure comparability, we standardize how texts are segmented. We use a two-level segmentation strategy. First, each document is split into individual \emph{sentences} using Stanza NLP toolkit \citep{QiEtAl20}. This yields a consistent base unit across all languages, since sentence boundaries are generally well-defined and preserved in translation.

Next, we group sentences into single-topic chunks (\emph{pseudo-paragraphs}) to create the segments that we would embed. For that purpose, we use linearly penalized segmentation (\emph{PELT}) algorithm by \citet{KillickEtAl12}. First, for each sentence in the original-language document we compute a sentence embedding using the language-specific sentence transformer model chosen as the segmentation model (see Appendix 1 for a list of models). We thus obtain for each document a sequence of $n$ sentence embeddings $\{\mathbf{x}_i\}_{i=1}^n \in \left(\mathbb{R}^d\right)^n$. For each sequence, we seek a set of change-points $\{\tau_k\}_{k=1}^K$ that partitions the sequence (and thus the document) into $K+1$ contiguous segments, so as to minimize a penalized within-segment cost given, for any segment $[a,b]$, by the empirical RKHS variance under a Gaussian (RBF) kernel:
\begin{equation*}
C(a,b) = (b - a + 1) - \frac{1}{b - a + 1}
  \sum_{i=a}^b \sum_{j=a}^b
  \exp\!\left(-\frac{\|\mathbf{x}_i - \mathbf{x}_j\|^2}{2\sigma^2}\right),
\end{equation*}
which is small when the embeddings within the interval are tightly clustered in the kernel feature space and large when they are dispersed. We then solve
\begin{equation*}
\min_{0=\tau_0<\tau_1<\cdots<\tau_K<\tau_{K+1}=n}
\quad
\sum_{k=0}^K C(\tau_k+1,\tau_{k+1}) + \beta K,
\end{equation*}
where $\beta > 0$ is a penalty that discourages over-segmentation, subject to the additional constraint that no segment includes fewer sentences than $\lfloor n / 20 \rfloor$ (ensuring enough context within each segment to capture a coherent idea). The kernel bandwidth $\sigma^2$ is set by the median heuristic over pairwise distances within the document. To find this global minimum efficiently, the PELT algorithm uses dynamic programming with a pruning rule.

We use PELT implementation included in the \texttt{ruptures} package for Python \citep{TruongEtAl18,TruongEtAl20}. We use penalization constant $1$, except that for documents where this approach leads to creation of only one pseudo-paragraph we instead use penalization constant $0.5$.

There remains a problem of segmenting translated texts, as there need not exist any bijective mapping between original-language sentences and post-translation sentences. We cannot segment translated texts independently, since in the absence of an assumption that translation preserves semantic similarity -- and this is the very claim we are attempting to verify -- there is no guarantee that translation-based segmentation will be similar to the original segmentation. Translating individual sentences would also make little sense, as it would involve losing contextual information that contributes to translation quality. We solve that problem by recovering alignment between original and translated sentences. Most of the sentence alignment algorithms are designed to work with unordered corpora \citep[see, e.g.,][]{GaleChurch93,Moore02a}. However, we may assume that translation does not alter sentence ordering, and therefore that ordering information is available. Accordingly, we use a custom monotone sequence-to-sequence mapping algorithm.

First, we convert each sentence of the source document and its English translation into length-normalized embeddings with a multilingual \emph{LaBSE} model \citep{FengEtAl22}. Given two ordered sequences of embeddings, we treat sentence alignment as a restricted sequence-to-sequence mapping problem and solve it with a monotone dynamic-programming algorithm equivalent to the Needleman–Wunsch global aligner \cite{NeedlemanWunsch70} but scored by semantic similarity rather than edit distance. Diagonal moves produce one-to-one links, while horizontal or vertical moves allow contiguous many-to-one (split/merge) relations; a negative gap penalty discourages but does not forbid such moves, thus yielding the optimal monotonic path that maximizes total cosine similarity subject to the contiguity constraint.

Note that our segmentation procedure constrains target-text boundaries to align with source-text boundaries via LaBSE-based monotone sentence alignment. Independent target-side segmentation would assume that meaningful segment boundaries are themselves preserved under translation, which is a stronger assumption than the one we test (preservation of pairwise similarity relationships among aligned segments). However, at least formally speaking, our findings characterize translation invariance \emph{conditional on a specific cross-lingual alignment procedure}, and a different aligner (e.g., one based on syntactic rather than embedding-based correspondence) could in principle yield different verdicts. On the other hand, alignment quality affects all pipeline types symmetrically -- both pre-translation and post-translation embeddings are computed on segments derived from the same alignment -- so the relative comparison among pipeline types is robust to alignment-induced noise.

\subsection{Pooling} \label{sec:pooling}

Transformer-based models have a fixed maximum context length (typically 512 to 4096 tokens). Encoding entire pseudo-paragraphs at once can exceed this limit, forcing truncation, which inevitably results in incomplete embeddings. Accordingly, we compute embeddings for sentences rather than paragraphs, and then aggregate them using \emph{centrality-weighted pooling}: a paragraph embedding is a weighted average of sentence embeddings, where the weight of each sentence reflects its ``importance'' in the document’s overall semantic structure. Formally, that importance is operationalized in terms of PageRank \citep{BrinPage98} centrality score in a fully connected sentence graph with edge weights equal to cosine similarity of the endpoint sentences. Because PageRank emphasizes sentences that are well connected to many others, centrality-weighted pooling captures the core semantic content of the paragraph in a fully unsupervised, parameter-free manner, while naturally down-weighting peripheral or outlier sentences\footnote{Because cosine similarity is bounded in $[-1,1]$ and PageRank requires non-negative edge weights, we truncate negative similarities to zero before constructing the graph. In practice this affected $<1\%$ of edges in our corpus owing to the well-documented anisotropy of transformer-based sentence embeddings \citep{Ethayarajh19}}.

\subsection{Similarity Measurement}

Having computed paragraph embeddings, for every language and every model we construct a paragraph similarity matrix. As our measure of text similarity we use \emph{cosine similarity}, given by
\begin{equation*}
    S(\mathbf{x}, \mathbf{y}) = \frac{\mathbf{x} \cdot \mathbf{y}}{\lVert\mathbf{x}\rVert_2 \lVert\mathbf{y}\rVert_2} = \frac{\sum_{i=1}^{d} x_i y_i}{\sqrt{\sum_{i=1}^{d} x_i^2} \sqrt{\sum_{i=1}^{d} y_i^2}},
\end{equation*}
where $\mathbf{x}$ and $\mathbf{y}$ are the embedding vectors, $d$ is the dimension of the embedding space, $\lVert \cdot \rVert_2$ denotes the $L_2$ (Euclidean) norm, and $\cdot$~denotes the inner product. The result, for every language-model pair, is an $N \times N$ similarity matrix, where $N$ is the number of pseudo-paragraphs in the dataset. Entry $(i,j)$ in this matrix is the cosine similarity between the embedding of paragraph $i$ and paragraph $j$.

We generate distinct matrices for original-language texts and for translated texts. Since ordering of paragraphs is constant, the cell $(i,j)$ in the original language matrix corresponds to the exact same pair of content segments as $(i,j)$ in the translated matrix. Crucially, we do not mix languages when computing similarities – each matrix is intralingual, built either on the original texts in language $L$ or on their English translations. The similarity matrix describes the semantic structure of the corpus under a specific model. Any changes in this matrix after translation would indicate that translation might be altering semantic relationships. By contrast, if the matrices for original and translation look very similar, it suggests the semantic structure is invariant under translation (for that model). These matrices thus form the basis for our invariance analysis.

\subsection{Evaluation Metrics} \label{sec:evalMetrics}

To quantitatively assess invariance, we directly compare the original-language and translated similarity matrices using correlation analysis. For each language, we compute Pearson correlation coefficients between each pair of similarity matrices. In practice, since each matrix is symmetric, we flatten only the upper-triangular entries to form two vectors of similarity scores. This correlation-based approach is essentially a form of \emph{Representational Similarity Analysis} (RSA) applied cross-lingually \cite{BeinbornChoenni20}. It treats the similarity matrix as a representation of the paragraph’s semantic structure and asks whether such representations for translated texts differ from those for original texts. A high Pearson correlation (near $1$) indicates that the semantic distances between all pairs of paragraphs in the original text are very closely preserved after translation. In other words, the model ``sees'' the relationships between paragraphs the same way in the source language and in English. Conversely, a lower correlation would reveal discrepancies -- certain paragraph pairs that were similar in the original may no longer appear as similar after translation, or vice versa, signaling a semantic shift or noise introduced by translation. Prior work on multilingual embeddings suggests that if a model truly captures language-agnostic semantics, the relative similarities should be unchanged by translation \cite{VasilyevEtAl24}. In fact, multilingual sentence embedding models are often trained explicitly to enforce this property \cite{ArtetxeSchwenk19,ReimersGurevych20}.

There are, however, two challenges involved here. First, since language models do not perfectly capture the semantic structure of a text, even for same-language models the correlation between similarity matrices will not be perfect. Hence, we need to distinguish semantic shift introduced by translation from random noise traceable to imperfect representation by the language models involved. Second, correlations enable us to compare how machine translation performs across languages and methods, but they do not provide an intuitively understandable measure of absolute performance. For instance, it is difficult to tell whether correlation $0.8$ between original and translated similarity matrices is good or unsatisfactory.

Use of multiple embedding models enables us to address those two problems. We compute correlations not only between original-language models and English-language models working on translated texts, but also between models within each of those classes. However, the original-language model class is heterogeneous in architecture and training quality: high-resource languages benefit from sentence-encoder architectures specifically optimized for semantic similarity, while lower-resource languages must rely on contextual word embedding models with mean pooling. The English models used post-translation, by contrast, are uniformly state-of-the-art sentence encoders selected from English-specific benchmarks. Comparing the original-language pipeline directly against the translation pipeline therefore conflates two distinct effects: translation-induced semantic distortion and the shift from a heterogeneous original-language model class to a homogeneous, performance-optimized English class. To disentangle these effects, we also analyze how translation affects similarities under multilingual models, which constitute a homogeneous family of architectures applied to both original and translated text. A finding of widespread invariance under this hypothesis, contrasted with the more variable verdicts under the baseline and best-model hypotheses, would indicate that a meaningful share of the apparent translation effect is in fact attributable to the architectural asymmetry rather than to translation per se.

For the multilingual hypothesis, we strengthen the architectural-control argument by restricting the cross-version comparison to \emph{same-model pairs}: that is, we compare each multilingual model on original-language text against \emph{the same multilingual model} applied to the corresponding translated text, rather than allowing any multilingual model on original text to be compared against any (possibly different) multilingual model on translated text. This restriction holds the encoder identity exactly constant across the translation step, so the candidate class measures only the perturbation induced by translation while the reference class continues to capture the natural disagreement among different multilingual models on original-language text. Sample size for the candidate class is correspondingly smaller (one pair per multilingual model per language, rather than one pair per ordered model pair), which we account for in the inferential procedure described below.

Accordingly, for each language, we test four hypotheses expressed in terms of stochastic orderings:
\begin{itemize}
\item \textbf{baseline invariance hypothesis} (\emph{semantic similarity is invariant under translation for a random language model}): \\
the correlation between the similarity matrices for two randomly chosen original-language models is \emph{not greater} than the correlation between similarity matrices for (i) a randomly chosen original-language model and (ii) a randomly chosen post-translation English model;
\item \textbf{best-model invariance hypothesis} (\emph{semantic similarity is invariant under translation for the best-performing post-translation model}): \\
the correlation between the similarity matrices for two randomly chosen original-language models is \emph{not greater} than the correlation between similarity matrices for (i) a randomly chosen original-language model and (ii) the best-performing post-translation English model, where ``best'' is defined by the highest average correlation with original-language similarity matrices;
\item \textbf{multilingual invariance hypothesis} (\emph{semantic similarity is invariant under translation for a random multilingual model}): \\
the correlation between the similarity matrices of two randomly chosen multilingual models operating on original-language data is \emph{not greater} than the correlation between similarity matrices for (i) a randomly chosen multilingual model operating on original-language data and (ii) \emph{the same multilingual model} operating on translated data;
\item \textbf{performance equivalence of post-translation vs. multilingual embeddings} (\emph{mono-lingual embeddings of translated texts preserve semantic similarity no worse than multilingual embeddings}): \\
the correlation between (i) a randomly chosen original-language model and (ii) a randomly chosen multilingual model applied to the original texts is \emph{not greater} than the correlation between (i) a randomly chosen original-language model and (ii) a randomly chosen post-translation English model.
\end{itemize}

For each of our four hypotheses, we cannot rely on a conventional null-hypothesis test of $\mu_A = \mu_B$, because failure to reject such a null does not constitute positive evidence of invariance. Substantively, all four hypotheses are concerned with deviations in only one direction: translation distortion that depresses similarity is undesirable, but translation that incidentally tightens similarity relative to the original is not a violation of any claim we make. We therefore adopt the one-sided non-inferiority framing \cite{Schuirmann87,Lakens17}, which inverts the inferential burden by treating inferiority as the null. For each language $L$ and each pair of pipeline types $(A, B)$ under comparison, we test
$$
H_0: \mu_A^L - \mu_B^L \geq \delta_L
$$
against the alternative $H_1: \mu_A^L - \mu_B^L < \delta_L$, where $\delta_L > 0$ is a pre-specified invariance margin and $A$ is the reference pipeline type whose performance $B$ is being compared against. Rejection at level $\alpha$ establishes that $B$ is equivalent to $A$ within margin $\delta_L$, which is operationally equivalent to verifying that the upper end of a $(1-2\alpha)$-confidence interval for $\mu_A^L - \mu_B^L$ falls below $\delta_L$. Failure to reject the inferiority null does not by itself imply non-equivalence; we additionally classify a language as exhibiting demonstrated translation distortion when the lower end of the same confidence interval exceeds $\delta_L$, and as indeterminate otherwise. This three-way partition is methodologically superior to the binary reject/fail-to-reject framework, because it explicitly distinguishes languages where the data positively support invariance from those where the available evidence is simply insufficient.

The invariance margin is calibrated to each language's intrinsic model-disagreement scale, $\delta_L := \kappa \cdot \sigma_L$, where $\kappa \in \mathbb{R}_{+}$ and $\sigma_L := \sqrt{2\hat{\sigma}^2_{\mathrm{cfg},L} + \hat{\sigma}^2_{\mathrm{resid},L}}$ is the model-based estimate of the marginal standard deviation of a correlation in the reference class for the hypothesis at hand (original language model pairs for the baseline, best-model, and performance-equivalence hypotheses; multilingual model pairs for the multilingual hypothesis), derived from a crossed multi-membership random-effects fit \cite{BrowneEtAl01} that pools across all reference-class pairs in language $L$. We report our primary results for $\kappa = 1$, which corresponds to the invariance margin being equal to the typical within-reference-class heterogeneity --- so translation is deemed invariant when its induced perturbation is no larger than that arising from natural disagreement among reference-class model pairs --- and report sensitivity at $\kappa \in \{0.5, 1.5, 2\}$. Confidence intervals for $\mu_A^L - \mu_B^L$ are computed using the same multi-membership specification with pair-type as a fixed effect, which honors the dyadic dependence structure (each correlation involves two embedding configurations, both of which appear in many other correlations) that a naive treatment of the matrix entries as independent observations would ignore; for the best-model hypothesis, where the candidate $B$ is itself selected from the data as the post-translation model with the highest mean correlation against $A$, we instead use a configuration-level cluster bootstrap that re-selects the best candidate within each replicate, propagating the model-selection uncertainty into the resulting confidence interval. Because the reported decisions involve simultaneous claims across all 28 languages, we additionally apply the Benjamini-Hochberg procedure \cite{BenjaminiHochberg95} to control the false-discovery rate of invariant and distorted declarations separately, at level $q = 0.05$, with a corresponding adjustment to language-level decisions reported alongside the uncorrected results.

\section{Impact on Downstream Tasks}

There remains yet another question: does the translation-induced semantic shift impact downstream tasks? To answer that question, we also evaluate how machine translation affects three common downstream tasks that depend on similarity measures: \emph{classification}, \emph{clustering}, and \emph{dimensionality reduction}.

\subsubsection{Classification}

As a test classification problem, we have chosen ideological classification: assignment of parties to party families (conservative, liberal, social-democratic, Christian Democratic, green, nationalist, etc.) on the basis of manifesto fragments. Our label data source is the Manifesto Research Project Dataset \cite{LehmannEtAl23a}, a companion dataset to the Manifesto Corpus, where each party is assigned to one of 12 party families. We use a random forest classifier \cite{Breiman01}.

To evaluate classifier performance, we use macro $F1$ scores \citep{OpitzBurst21} and Matthews Correlation Coefficient \citep{ChiccoJurman20}. We compare distributions of both metrics for original-language, multilingual, and post-translation models. To measure agreement between classifiers run under different embedding models and text versions, we use the adjusted Rand index \citep{Rand71,HubertArabie85}, for which we run the same statistical tests as for correlations between similarity matrices.

\subsubsection{Clustering}

To evaluate the impact of machine translation on downstream clustering tasks, we first run spherical $k$-means clustering (with $n = 20$) on the embedding space generated by each model for each text version (original / translated). Then, we measure agreement between partitions generated under different embedding models and text versions using the adjusted Rand index \citep{Rand71,HubertArabie85}, and test similarity hypotheses as in the case of classification tasks.

\subsubsection{Dimensionality reduction}

To assess the impact of machine translation on downstream clustering tasks, we use Uniform Manifold Approximation and Projection (UMAP) algorithm \citep{McInnesEtAl18,McInnesEtAl20} to reduce dimensionality of the embeddings. To preserve local density information, we employ the densMAP variant of the original UMAP \citep{NarayanEtAl21}. We assume the topology of the original embedding space to be the metric topology induced by the cosine metric, fix the topological dimension of the projection space at $16$, and fix the local neighborhood size hyperparameter at $32$. Then, we measure similarities between low-dimensional embeddings, calculate correlation coefficients, and test hypotheses as in the case of unreduced embeddings.

\section{Results}

\begin{figure*}[t]
    \includegraphics[width=\textwidth]{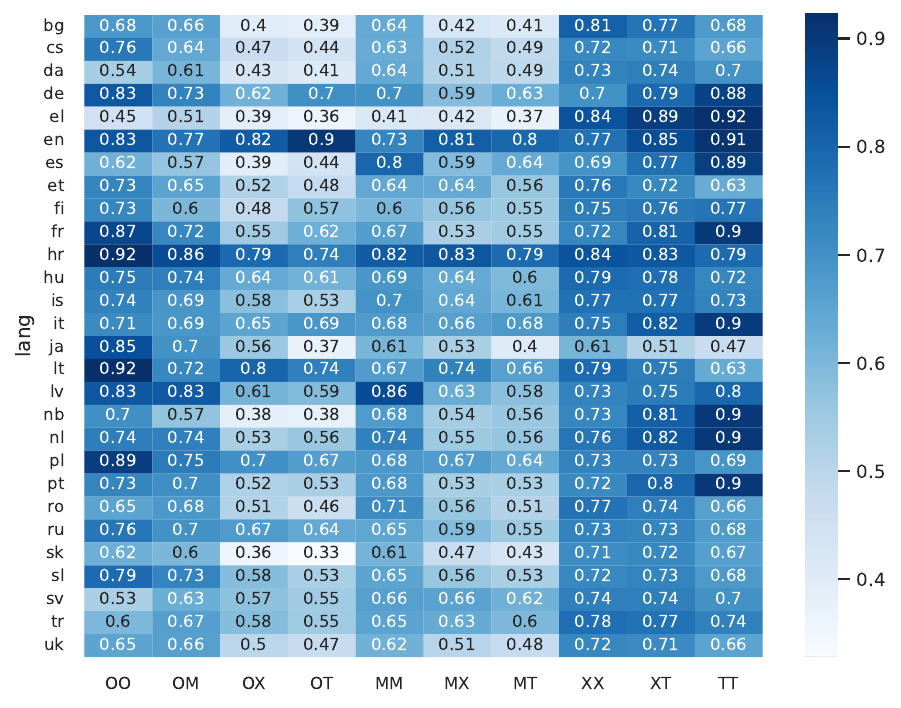}
    \caption{Average correlations by language and by the type of pipeline used to generate the embeddings.}
    \label{fig:heatmap}
\end{figure*}

We begin our discussion of the results with some descriptive statistics. Each similarity matrix is assigned to one of four pipeline types: \textbf{O}~--~original-language model embedding, \textbf{T} -- translation + English monolingual model embedding, \textbf{M} -- multilingual model embedding, and \textbf{X} -- translation + multilingual model embedding. It follows that each correlation falls within one of ten types: OO, TT, MM, XX, OT, OM, OX, MT, MX, XT, depending on the types of both models involved. For each language and each of those ten types, we first report average correlation coefficients, see Figure \ref{fig:heatmap}.

At first glance, it appears that agreement between O and T pipelines is lower on the average than agreement between O and O pipelines, or between O and M pipelines (Swedish is the sole exception here). At the same time, the agreement between M and X pipelines is likewise lower than between M and M pipelines, though the difference in this case is smaller. These observations strongly suggest that translation acts as a systematic perturbation to the agreement structure, an effect that we quantify more rigorously next.

That said, languages differ markedly in how robust they are to translation: for some the semantic similarity shift is small (average correlation between similarity matrices on the order of $0.8$ to $0.9$), while for others it is significant (average correlation between matrices below $0.5$). No clear pattern is evident here: susceptibility to semantic similarity shift appears to be independent of whether the original language is a high- or low-resource one, and of the language family.

More formal statistical analysis presents a more nuanced view. For each language and each of the four hypotheses identified above, we test the invariance null using the procedure described in Subsection~\ref{sec:evalMetrics}, with confidence intervals derived from the multi-membership crossed random-effects model that respects the dyadic dependence among correlations sharing a common embedding configuration; for the best-model hypothesis, where the candidate pipeline is selected from the data, we instead use a configuration-level cluster bootstrap that re-selects the best candidate within each replicate. The invariance margin for each language is set to $\delta_L = \kappa \cdot \sigma_L$, where $\sigma_L$ is the model-based estimate of the marginal standard deviation of a correlation in the hypothesis's reference class ($OO$ for the baseline, best-model, and performance-equivalence hypotheses; $MM$ for the multilingual hypothesis) and $\kappa = 1$ is our primary tolerance. To control the false-discovery rate across the $28$ languages within each hypothesis, we apply the Benjamini-Hochberg procedure at $q = 0.05$, separately to the invariance $p$-values and to the corresponding distortion $p$-values. For the three directional hypotheses (baseline, best-model, multilingual) we classify each language as \emph{invariant} when the upper end of the $90\%$ confidence interval falls below $\delta_L$ (equivalently, when $B$ correlation is non-inferior to $A$ within tolerance), \emph{distorted} when the lower end exceeds $\delta_L$, and \emph{indeterminate} when the interval straddles $\delta_L$. The performance-equivalence hypothesis (OM--OT) is genuinely two-sided, since both directions of deviation matter substantively, and accordingly produces a four-way decision: \emph{equivalent} (CI within $(-\delta_L, +\delta_L)$), \emph{OT inferior} (CI lower end exceeds $+\delta_L$, OT detectably worse than OM), \emph{OT superior} (CI upper end below $-\delta_L$, OT detectably better than OM), and \emph{indeterminate}.

The aggregate distribution of decisions is summarized in Table~\ref{tbl:sensitivity}; per-language results (effect magnitudes and test outcomes) appear in Table~\ref{tbl:allHypotheses}, while detailed results are reported in Appendix 2. Each language is characterized by the joint pattern of its four verdicts at $\kappa = 1$, and these joint patterns reveal a clear stratification of languages by how well translation-based pipelines work for them.

Five languages -- German, Italian, Portuguese, Spanish, and Ukrainian -- exhibit positive verdicts on all four hypotheses: translation preserves similarity (OO--OT invariant) on the average and under the best model, translation does not degrade multilingual encoders (MM--MX invariant), and translation-plus-monolingual is interchangeable with direct multilingual encoding (OM--OT equivalent). For these languages the choice of cross-lingual pipeline is essentially unconstrained: any of the three approaches we tested produces semantic similarity matrices that are statistically indistinguishable from the others, and from the original-language baseline. Five further languages -- Bulgarian, Greek, Hungarian, Icelandic, and Swedish -- exhibit positive verdicts on three of four hypotheses with the fourth merely indeterminate (and in no case distorted); for them translation pipelines are similarly safe, with the unresolved hypothesis reflecting limited statistical power rather than evidence of pipeline failure.

Four languages -- French, Japanese, Latvian, and Lithuanian -- show the opposite pattern: the baseline OO--OT and best-model OO--OT$_\text{best}$ hypotheses are decisively rejected, indicating that for these languages translation introduces semantic distortion beyond the tolerated margin even when the best post-translation embedding model is selected. For the two Baltic languages this distortion is so substantial that it also manifests in the OM--OT comparison: translation-plus-monolingual performs detectably worse than direct multilingual encoding, suggesting that the problem lies primarily with the eTranslation step rather than with the post-translation embedding model. For French and Japanese the OM--OT comparison is indeterminate, indicating that while translation produces measurable distortion, multilingual encoders applied to the original text do not clearly outperform the translation pipeline either. Multilingual invariance (MM--MX) holds for all four languages, confirming that the failure mode in this group is specifically the translation step combined with monolingual encoding, not multilingual encoding under translation.

English is the unique outlier in the opposite direction: the OM--OT comparison places it as OT \emph{superior} to OM, meaning that running state-of-the-art monolingual English encoders on the (identity-preserving) eTranslation output beats the multilingual baselines at fine-grained semantic similarity. This finding is unsurprising on reflection, since English-to-English ``translation'' through eTranslation is near-identity, so for English the OT pipeline is effectively the strong monolingual encoder operating on minimally-modified text. We report this verdict for completeness but caution that English is methodologically not comparable to the other languages in the OM--OT comparison and that this result should not be aggregated with the substantively-translated cases.

The remaining thirteen languages are characterized predominantly by indeterminate verdicts, particularly on the baseline OO--OT comparison: the data are too sparse to support a positive claim of either invariance or distortion. This indeterminate category accounts for $38\%$ of language-hypothesis pairs at $\kappa = 1$ -- the methodologically honest acknowledgment that for many languages the available evidence cannot conclusively distinguish small-but-real translation distortions from sampling noise. Notably, all but Finnish nonetheless exhibit MM--MX invariance, indicating that multilingual encoders applied directly to translated text are robust even where the OO--OT comparison cannot resolve.

Under Benjamini-Hochberg adjustment at $q=0.05$, five borderline positive verdicts become indeterminate: Romanian and Russian on the best-model hypothesis, and Dutch, Portuguese, and Ukrainian on the OM-OT comparison. All four distortion verdicts on the baseline and best-model hypotheses, the three OT-inferior verdicts on OM-OT (Latvian, Lithuanian, Slovenian), and English's OT-superior verdict are unaffected, as is the widespread MM-MX invariance. The cluster structure described above therefore retains its rough shape, with Portuguese and Ukrainian moving from the all-positive cluster to the three-of-four cluster, and Romanian and Russian moving from three-of-four to predominantly-indeterminate.

Sensitivity analyses at $\kappa \in \{0.5, 1.5, 2\}$ (Table~\ref{tbl:sensitivity}) confirm the qualitative pattern. The four languages established as distorted on the baseline hypothesis at $\kappa = 1$ remain distorted at $\kappa = 0.5$, and three of them (French, Japanese, Lithuanian) remain distorted even under the most permissive $\kappa = 2$. Conversely, only four of the eight languages established as invariant at $\kappa = 1$ retain that classification under the strictest $\kappa = 0.5$, with the remainder migrating to indeterminate. The grouping of languages by joint verdict pattern is therefore robust at the level of the most-invariant and least-invariant clusters, with intermediate languages shifting between adjacent categories as $\kappa$ varies.

\begin{table}[h]
\centering
\small
\setlength{\tabcolsep}{4pt}
\begin{tabularx}{\columnwidth}{@{}Xcccc@{}}
\toprule
Language & OO--OT & OO--OT$_\text{best}$ & MM--MX & OM--OT \\
\midrule
Bulgarian & \textbf{.137} & \textbf{.056} & -.046 & \textbf{.130} \\
Croatian & .271 & .184 & \textbf{-.105} & .194 \\
Czech & .246 & .179 & \textbf{-.073} & .118 \\
Danish & .060 & \textbf{.013} & \textbf{-.074} & .132 \\
Dutch & .140 & .121 & \textbf{-.033} & .028 \\
English & -.054 & \textbf{-.080} & \textbf{-.366} & \textcolor{blue}{\textbf{-.265}} \\
Estonian & .192 & .154 & \textbf{-.165} & \textit{.156} \\
Finnish & .060 & \textbf{.031} & -.243 & \textbf{-.018} \\
French & \textcolor{red}{\textbf{.201}} & \textcolor{red}{\textbf{.175}} & \textbf{-.052} & -.056 \\
German & \textbf{.026} & \textbf{9.6$\mathbf{E}$-05} & \textbf{.033} & \textbf{9.6$\mathbf{E}$-03} \\
Greek & \textbf{.071} & \textbf{-.066} & .087 & \textbf{.146} \\
Hungarian & .100 & \textbf{.049} & \textbf{-.046} & \textit{\textbf{.094}} \\
Icelandic & .151 & \textbf{.057} & \textbf{-.114} & \textbf{.105} \\
Italian & \textbf{1.7$\mathbf{E}$-03} & \textbf{-.034} & \textbf{-.177} & \textbf{-.098} \\
Japanese & \textcolor{red}{\textbf{.423}} & \textcolor{red}{\textbf{.171}} & \textbf{-.203} & .200 \\
Latvian & \textcolor{red}{\textbf{.229}} & \textcolor{red}{\textbf{.132}} & \textbf{-.178} & \textcolor{red}{\textbf{.119}} \\
Lithuanian & \textcolor{red}{\textbf{.352}} & \textcolor{red}{\textbf{.275}} & \textbf{-.100} & \textcolor{red}{\textbf{.206}} \\
Norwegian & .304 & .285 & \textbf{-.031} & .208 \\
Polish & .139 & .017 & \textbf{-.220} & -.086 \\
Portuguese & \textbf{.169} & \textbf{.131} & \textbf{.013} & \textbf{.123} \\
Romanian & .086 & \textbf{-.019} & \textbf{-.134} & .085 \\
Russian & .062 & \textbf{-3.5$\mathbf{E}$-03} & \textbf{-.092} & -.066 \\
Slovak & .305 & .120 & \textbf{-5.9$\mathbf{E}$-03} & \textit{.317} \\
Slovenian & .260 & .183 & \textbf{-.096} & .220 \\
Spanish & \textbf{.161} & \textbf{.110} & \textbf{-.110} & \textit{\textbf{.137}} \\
Swedish & \textbf{-.089} & \textbf{-.143} & -.252 & \textbf{-.014} \\
Turkish & 5.6$\mathrm{E}$-03 & \textbf{-.048} & \textbf{-.135} & .054 \\
Ukrainian & \textbf{.031} & \textbf{-.018} & \textbf{-.089} & \textbf{.020} \\
\bottomrule
\end{tabularx}
\caption{Per-language estimates $\hat D = \hat\mu_A - \hat\mu_B$ for the four invariance hypotheses at $\kappa = 1$. \textbf{Bold black}: invariance / equivalence. \textcolor{red}{\textbf{Red bold}}: the translation pipeline exhibits greater distortion than the reference pipeline. \textcolor{blue}{\textbf{Blue bold}}: the translation pipeline exhibits lesser distortion than the reference pipeline (OM--OT only). Plain: indeterminate. \textit{Italicized} OM--OT entries indicate languages for which the mixed-model fit failed to converge and the reported estimate is from the cluster-bootstrap backend instead.}
\label{tbl:allHypotheses}
\end{table}

\begin{table}[h]
\centering
\small
\setlength{\tabcolsep}{4pt}
\begin{tabularx}{\columnwidth}{@{}Xcrrrr@{}}
\toprule
hypothesis & $\kappa$ & invariant & indeterm. & distort. & super. \\
\midrule
Baseline OO-OT & 0.5 & 4 & 16 & 8 & \\
 & \textbf{1.0} & \textbf{8} & \textbf{16} & \textbf{4} & \\
 & 1.5 & 11 & 14 & 3 & \\
 & 2.0 & 13 & 12 & 3 & \\
\midrule
Best model OO-OT$_\text{best}$ & 0.5 & 9 & 13 & 6 & \\
 & \textbf{1.0} & \textbf{16} & \textbf{8} & \textbf{4} & \\
 & 1.5 & 18 & 6 & 4 & \\
 & 2.0 & 21 & 3 & 4 & \\
\midrule
Multilingual MM-MX & 0.5 & 19 & 9 & 0 & \\
 & \textbf{1.0} & \textbf{24} & \textbf{4} & \textbf{0} & \\
 & 1.5 & 27 & 1 & 0 & \\
 & 2.0 & 27 & 1 & 0 & \\
\midrule
OM-OT perf. equiv. & 0.5 & 2 & 18 & 7 & 1 \\
 & \textbf{1.0} & \textbf{9} & \textbf{15} & \textbf{3} & \textbf{1} \\
 & 1.5 & 12 & 15 & 1 & 0 \\
 & 2.0 & 18 & 9 & 1 & 0 \\
\bottomrule
\end{tabularx}
\caption{Decision counts across all 28 languages for each hypothesis at four invariance-margin tolerances $\kappa$. The primary specification ($\kappa = 1$) is highlighted in bold. For the three directional hypotheses (top three blocks), decisions are \emph{invariant} / \emph{indeterminate} / \emph{distorted}. For the two-sided OM-OT hypothesis (bottom block), decisions are \emph{equivalent} / \emph{indeterminate} / \emph{OT inferior} / \emph{OT superior}; verdicts come from the mixed-model fit for 24 languages and from cluster-bootstrap inference for the four (Estonian, Hungarian, Slovak, Spanish) for which the mixed-model fit failed to converge.}
\label{tbl:sensitivity}
\end{table}

\subsection{Impact of translation on downstream tasks}

The formal hypothesis-test framework introduced in Subsection~\ref{sec:evalMetrics} extends naturally to downstream tasks by replacing similarity-matrix correlations with adjusted Rand index between classifier-pair or clustering-pair partitions, or with correlations between UMAP-reduced similarity matrices. Tables~\ref{tbl:allHypothesesClass}--\ref{tbl:sensitivityUmap} report the per-language verdicts and decision counts for each of the three downstream tasks.

The classification analysis (Table~\ref{tbl:allHypothesesClass}, Table~\ref{tbl:sensitivityClass}) finds no language with established distortion on either the baseline OO--OT or the best-model OO--OT$_\text{best}$ hypothesis at $\kappa = 1$: the four languages flagged as distorted in the similarity analysis (French, Japanese, Latvian, and Lithuanian) all migrate to either invariant or indeterminate verdicts when the unit of analysis shifts from embedding-similarity matrices to classifier predictions. Slovenian alone retains an OT-inferior verdict on the OM--OT comparison, and English no longer registers as OT superior -- its translation-pipeline classifier is statistically indistinguishable from the multilingual baseline. Sensitivity analysis at $\kappa = 0.5$ produces 4 newly flagged distorted languages on the baseline hypothesis, but none of these survives at $\kappa = 1$ or above.

We have also found no impact of translation on classification performance: in terms of $F_1$ scores, the translation pipeline yielded performance ($.388$) slightly worse than the original-language pipeline ($.393$), though better than the multilingual pipeline ($.378$), but the differences were not statistically significant. In terms of MCC, it actually outperformed all the other pipelines ($.397$ vs. $.376$ for original-language), but again the differences were not significant.

The clustering analysis (Table~\ref{tbl:allHypothesesClus}, Table~\ref{tbl:sensitivityClus}) occupies an intermediate position. At $\kappa = 1$, two languages (French and Japanese) retain distortion verdicts on the baseline OO--OT hypothesis. French additionally exhibits distortion on the best-model hypothesis at all four $\kappa$ values, indicating particularly persistent translation-induced distortion of cluster structure for that language; Japanese, by contrast, is rescued at the best-model level (best-T clustering matches the typical O-clustering), suggesting that the choice of post-translation embedding model matters more for Japanese than for French. Japanese alone retains an OT-inferior verdict on the OM--OT comparison, while English remains OT superior.

The UMAP analysis (Table~\ref{tbl:allHypothesesUmap}, Table~\ref{tbl:sensitivityUmap}) shows the noisiest pattern of any downstream task. UMAP's stochastic optimization inflates within-class variance: $\sigma_{OO}$ values for UMAP-reduced similarity matrices range up to $1.77$ (Swedish) and $1.17$ (Russian), against $\sigma_{OO}$ typically below $0.5$ for the unreduced similarity analysis. Wider reference distributions translate to wider invariance margins and correspondingly inflated indeterminate counts: only 13 languages are invariant on the baseline hypothesis at $\kappa = 1$, with 14 indeterminate. Despite this noise, three substantively-meaningful verdicts persist: Latvian remains \emph{distorted} on baseline OO--OT and best-model OO--OT$_\text{best}$, and OT-inferior on OM--OT -- the strongest evidence in any analysis that UMAP applied to translated Latvian text degrades the original similarity structure. Czech is distorted on the best-model hypothesis as well, suggesting that no post-translation embedding model adequately preserves Czech document structure under UMAP reduction. Japanese flags OT-inferior on OM--OT with an effect size ($\hat{D} = 0.32$) larger than in any other analysis, indicating that UMAP amplifies rather than smooths translation-induced distortion for Japanese.

Across the downstream tasks, two patterns emerge. First, the four languages identified as distorted on the similarity-matrix baseline hypothesis (French, Japanese, Latvian, and Lithuanian) progressively dissolve as the analysis moves further from the raw embedding geometry. Classification at $\kappa = 1$ flags no language as distorted on either the baseline or best-model hypothesis; clustering retains French and Japanese; UMAP retains Latvian alone. The OT-superior verdict for English on OM--OT, present in the similarity and clustering analyses, similarly disappears under classification. This stratification confirms that translation-induced perturbations to the geometric structure of embeddings do not propagate uniformly to downstream task behavior. Tasks whose outputs are coarse, discrete decisions (12-way classification) are substantially less sensitive to translation distortion than tasks operating directly on embedding geometry; tasks producing partitions (clustering) lie between.

Second, multilingual invariance MM--MX holds robustly across all three downstream tasks: $22$ of $28$ languages invariant for classification, $26$ for clustering, and $24$ for UMAP, with the remaining languages indeterminate in each case. Across all 84 language $\times$ downstream-task combinations, no language exhibits distortion of multilingual encoders applied to translated text relative to multilingual encoders on original text. This robustness of multilingual encoding under translation -- consistent across all four analyses including the original similarity-matrix one ($24$ of $28$ invariant) -- is the most strongly-supported substantive conclusion of our study.

\begin{table}[t]
\centering
\small
\setlength{\tabcolsep}{4pt}
\begin{tabularx}{\columnwidth}{@{}Xcccc@{}}
\toprule
Language & OO-OT & OO-OT$_\text{best}$ & MM-MX & OM-OT \\
\midrule
Bulgarian & \textbf{.086} & \textbf{.071} & \textbf{.079} & \textbf{.089} \\
Croatian & .055 & .024 & \textbf{-.028} & .046 \\
Czech & .071 & .053 & \textbf{-4.8$\mathbf{E}$-03} & .057 \\
Danish & .037 & \textbf{.027} & 3.6$\mathrm{E}$-03 & .051 \\
Dutch & .034 & .022 & \textbf{9.9$\mathbf{E}$-03} & .031 \\
English & -.015 & \textbf{-.040} & \textbf{-.119} & -.043 \\
Estonian & .057 & .020 & \textbf{-.028} & \textbf{.052} \\
Finnish & 4.0$\mathrm{E}$-03 & \textbf{-.014} & -.029 & 1.2$\mathrm{E}$-03 \\
French & .071 & \textbf{.048} & \textbf{.013} & \textbf{.036} \\
German & \textbf{.019} & \textbf{-3.1$\mathbf{E}$-03} & \textbf{-.043} & \textbf{-.011} \\
Greek & .051 & \textbf{-.018} & \textbf{-.045} & .037 \\
Hungarian & .013 & \textbf{-.018} & \textbf{-.013} & \textit{\textbf{.025}} \\
Icelandic & \textbf{5.5$\mathbf{E}$-03} & \textbf{-.027} & -.019 & \textbf{.023} \\
Italian & \textbf{.034} & .018 & \textbf{-.044} & 7.4$\mathrm{E}$-03 \\
Japanese & \textbf{.011} & \textbf{-.018} & \textbf{9.8$\mathbf{E}$-03} & \textbf{-.023} \\
Latvian & .079 & .048 & -.080 & \textbf{-.011} \\
Lithuanian & .069 & \textbf{.052} & \textbf{-.035} & \textbf{.056} \\
Norwegian & .067 & .034 & \textbf{2.9$\mathbf{E}$-03} & .021 \\
Polish & \textbf{.023} & \textbf{3.4$\mathbf{E}$-04} & \textbf{-.040} & \textbf{-7.7$\mathbf{E}$-03} \\
Portuguese & .031 & \textbf{-7.3$\mathbf{E}$-03} & .022 & .036 \\
Romanian & \textbf{-.032} & \textbf{-.052} & \textbf{-.052} & \textbf{6.8$\mathbf{E}$-03} \\
Russian & .024 & \textbf{-.040} & \textbf{-.021} & \textbf{.010} \\
Slovak & .031 & \textbf{5.9$\mathbf{E}$-04} & \textbf{-.011} & \textit{\textbf{.044}} \\
Slovenian & .106 & .056 & \textbf{.039} & \textcolor{red}{\textbf{.083}} \\
Spanish & .014 & 3.4$\mathrm{E}$-03 & -5.3$\mathrm{E}$-03 & \textit{.024} \\
Swedish & \textbf{-.024} & \textbf{-.043} & \textbf{-.059} & \textbf{-9.0$\mathbf{E}$-03} \\
Turkish & \textbf{-8.6$\mathbf{E}$-03} & \textbf{-.040} & \textbf{-.055} & 1.3$\mathrm{E}$-03 \\
Ukrainian & .015 & \textbf{-.015} & \textbf{5.2$\mathbf{E}$-03} & .025 \\
\bottomrule
\end{tabularx}
\caption{Per-language estimates $\hat D = \hat\mu_A - \hat\mu_B$ on model-pair adjusted Rand index for the four invariance hypotheses at $\kappa = 1$. \textbf{Bold black}: invariance / equivalence. \textcolor{red}{\textbf{Red bold}}: the translation pipeline exhibits greater distortion than the reference pipeline. \textcolor{blue}{\textbf{Blue bold}}: the translation pipeline exhibits lesser distortion than the reference pipeline (OM--OT only). Plain: indeterminate. \textit{Italicized} OM-OT entries indicate languages for which the mixed-model fit failed to converge and the reported estimate is from the cluster-bootstrap backend instead.}
\label{tbl:allHypothesesClass}
\end{table}

\begin{table}[t]
\centering
\small
\setlength{\tabcolsep}{4pt}
\begin{tabularx}{\columnwidth}{@{}Xlrrrr@{}}
\toprule
hypothesis & $\kappa$ & invariant & indeterm. & distort. & super. \\
\midrule
Baseline OO-OT & 0.5 & 4 & 20 & 4 & \\
 & \textbf{1.0} & \textbf{9} & \textbf{19} & \textbf{0} & \\
 & 1.5 & 17 & 11 & 0 & \\
 & 2.0 & 23 & 5 & 0 & \\
\midrule
Best model OO-OT$_\text{best}$ & 0.5 & 13 & 15 & 0 & \\
 & \textbf{1.0} & \textbf{19} & \textbf{9} & \textbf{0} & \\
 & 1.5 & 24 & 4 & 0 & \\
 & 2.0 & 26 & 2 & 0 & \\
\midrule
Multilingual MM-MX & 0.5 & 16 & 12 & 0 & \\
 & \textbf{1.0} & \textbf{22} & \textbf{6} & \textbf{0} & \\
 & 1.5 & 24 & 4 & 0 & \\
 & 2.0 & 25 & 3 & 0 & \\
\midrule
OM-OT perf. equiv. & 0.5 & 10 & 17 & 1 & 0 \\
 & \textbf{1.0} & \textbf{20} & \textbf{7} & \textbf{1} & \textbf{0} \\
 & 1.5 & 25 & 3 & 0 & 0 \\
 & 2.0 & 26 & 2 & 0 & 0 \\
\bottomrule
\end{tabularx}
\caption{Decision counts for the classification-task analysis across all 28 languages for each hypothesis at four invariance-margin tolerances $\kappa$. The primary specification ($\kappa = 1$) is highlighted in bold. For the three directional hypotheses (top three blocks), decisions are \emph{invariant} / \emph{indeterminate} / \emph{distorted}. For the two-sided OM-OT hypothesis (bottom block), decisions are \emph{equivalent} / \emph{indeterminate} / \emph{OT inferior} / \emph{OT superior}; verdicts come from the mixed-model fit for 25 languages and from cluster-bootstrap inference for the three (Estonian, Hungarian, Slovak) for which the mixed-model fit failed to converge.}
\label{tbl:sensitivityClass}
\end{table}

\begin{table}[t]
\centering
\small
\setlength{\tabcolsep}{4pt}
\begin{tabularx}{\columnwidth}{@{}Xcccc@{}}
\toprule
Language & OO-OT & OO-OT$_\text{best}$ & MM-MX & OM-OT \\
\midrule
Bulgarian & \textbf{.061} & \textbf{.044} & \textbf{2.5$\mathbf{E}$-03} & \textbf{.040} \\
Croatian & \textbf{.012} & \textbf{-.030} & \textbf{-.120} & \textit{\textbf{-5.0$\boldsymbol{E}$-03}} \\
Czech & .113 & \textbf{.068} & \textbf{-.025} & \textbf{.041} \\
Danish & \textbf{9.7$\mathbf{E}$-03} & \textbf{-.011} & \textbf{-.048} & \textbf{.040} \\
Dutch & .206 & .182 & \textbf{.022} & .096 \\
English & -.048 & \textbf{-.071} & \textbf{-.672} & \textcolor{blue}{\textbf{-.172}} \\
Estonian & \textbf{-.061} & \textbf{-.092} & \textbf{-.119} & \textbf{-.043} \\
Finnish & \textbf{-.016} & \textbf{-.055} & \textbf{-.104} & -4.1$\mathrm{E}$-04 \\
French & \textcolor{red}{\textbf{.257}} & \textcolor{red}{\textbf{.239}} & \textbf{.024} & .082 \\
German & \textbf{.036} & \textbf{9.3$\mathbf{E}$-03} & \textbf{-.088} & \textbf{-.029} \\
Greek & .027 & \textbf{-.038} & \textbf{-.058} & -.011 \\
Hungarian & .093 & .031 & \textbf{-.030} & \textbf{.024} \\
Icelandic & \textbf{.026} & \textbf{4.2$\mathbf{E}$-04} & \textbf{-.051} & \textbf{4.6$\mathbf{E}$-03} \\
Italian & \textbf{-3.2$\mathbf{E}$-03} & \textbf{-.041} & -.096 & \textbf{-.055} \\
Japanese & \textcolor{red}{\textbf{.147}} & .043 & \textbf{.011} & \textcolor{red}{\textbf{.117}} \\
Latvian & .119 & -.039 & \textbf{-.081} & \textbf{-5.2$\mathbf{E}$-04} \\
Lithuanian & .082 & \textbf{.049} & \textbf{-.013} & \textbf{.044} \\
Norwegian & .128 & .109 & \textbf{.039} & .118 \\
Polish & .104 & .038 & \textbf{-.119} & -.049 \\
Portuguese & .099 & .086 & \textbf{.045} & .094 \\
Romanian & \textbf{-4.0$\mathbf{E}$-03} & \textbf{-.024} & \textbf{-.069} & \textbf{.025} \\
Russian & \textbf{.025} & \textbf{-9.5$\mathbf{E}$-03} & -.082 & \textbf{-.031} \\
Slovak & .093 & .073 & \textbf{-.012} & \textit{.088} \\
Slovenian & .105 & .063 & \textbf{-.019} & \textbf{.026} \\
Spanish & \textbf{.044} & \textbf{.034} & \textbf{-.029} & \textit{\textbf{.052}} \\
Swedish & \textbf{-.046} & \textbf{-.102} & \textbf{-.141} & -.028 \\
Turkish & \textbf{-.018} & \textbf{-.072} & \textbf{-.090} & \textbf{-6.5$\mathbf{E}$-03} \\
Ukrainian & \textbf{.018} & \textbf{-8.4$\mathbf{E}$-03} & \textbf{-.070} & -7.2$\mathrm{E}$-03 \\
\bottomrule
\end{tabularx}
\caption{Per-language estimates $\hat D = \hat\mu_A - \hat\mu_B$ on model-pair adjusted Rand index from the clustering analysis for the four invariance hypotheses at $\kappa = 1$. \textbf{Bold black}: invariance / equivalence. \textcolor{red}{\textbf{Red bold}}: the translation pipeline exhibits greater distortion than the reference pipeline. \textcolor{blue}{\textbf{Blue bold}}: the translation pipeline exhibits lesser distortion than the reference pipeline (OM--OT only). Plain: indeterminate. \textit{Italicized} OM-OT entries indicate languages for which the mixed-model fit failed to converge and the reported estimate is from the cluster-bootstrap backend instead.}
\label{tbl:allHypothesesClus}
\end{table}

\begin{table}[t]
\centering
\small
\setlength{\tabcolsep}{4pt}
\begin{tabularx}{\columnwidth}{@{}Xlrrrr@{}}
\toprule
hypothesis & $\kappa$ & invariant & indeterm. & distort. & super. \\
\midrule
Baseline OO-OT & 0.5 & 7 & 17 & 4 & \\
 & \textbf{1.0} & \textbf{14} & \textbf{12} & \textbf{2} & \\
 & 1.5 & 18 & 8 & 2 & \\
 & 2.0 & 18 & 9 & 1 & \\
\midrule
Best model OO-OT$_\text{best}$ & 0.5 & 15 & 12 & 1 & \\
 & \textbf{1.0} & \textbf{18} & \textbf{9} & \textbf{1} & \\
 & 1.5 & 22 & 5 & 1 & \\
 & 2.0 & 23 & 4 & 1 & \\
\midrule
Multilingual MM-MX & 0.5 & 16 & 12 & 0 & \\
 & \textbf{1.0} & \textbf{26} & \textbf{2} & \textbf{0} & \\
 & 1.5 & 26 & 2 & 0 & \\
 & 2.0 & 28 & 0 & 0 & \\
\midrule
OM-OT perf. equiv. & 0.5 & 6 & 19 & 2 & 1 \\
 & \textbf{1.0} & \textbf{16} & \textbf{10} & \textbf{1} & \textbf{1} \\
 & 1.5 & 19 & 9 & 0 & 0 \\
 & 2.0 & 19 & 9 & 0 & 0 \\
\bottomrule
\end{tabularx}
\caption{Decision counts for the clustering-task analysis across all 28 languages for each hypothesis at four invariance-margin tolerances $\kappa$. The primary specification ($\kappa = 1$) is highlighted in bold. For the three directional hypotheses (top three blocks), decisions are \emph{invariant} / \emph{indeterminate} / \emph{distorted}. For the two-sided OM-OT hypothesis (bottom block), decisions are \emph{equivalent} / \emph{indeterminate} / \emph{OT inferior} / \emph{OT superior}; verdicts come from the mixed-model fit for 25 languages and from cluster-bootstrap inference for the three (Spanish, Croatian, Slovak) for which the mixed-model fit failed to converge.}
\label{tbl:sensitivityClus}
\end{table}

\begin{table}[t]
\centering
\small
\setlength{\tabcolsep}{4pt}
\begin{tabularx}{\columnwidth}{@{}Xcccc@{}}
\toprule
Language & OO-OT & OO-OT$_\text{best}$ & MM-MX & OM-OT \\
\midrule
Bulgarian & \textbf{.096} & \textbf{-.076} & -.137 & \textbf{.029} \\
Croatian & .304 & .249 & \textbf{-.185} & \textbf{.140} \\
Czech & .556 & \textcolor{red}{\textbf{.509}} & \textbf{.126} & .423 \\
Danish & \textbf{.014} & \textbf{-.152} & \textbf{-.081} & \textbf{.066} \\
Dutch & .155 & .076 & \textbf{-.108} & \textbf{.014} \\
English & -.146 & \textbf{-.190} & \textbf{-.637} & -.297 \\
Estonian & .313 & .181 & \textbf{-.070} & \textit{.280} \\
Finnish & .175 & -.206 & \textbf{-.280} & .068 \\
French & \textbf{.167} & \textbf{.081} & \textbf{-.180} & \textbf{-8.0$\mathbf{E}$-03} \\
German & \textbf{.164} & \textbf{.134} & \textbf{.071} & \textbf{.194} \\
Greek & .223 & .070 & \textbf{.087} & .254 \\
Hungarian & \textbf{-.016} & \textbf{-.136} & \textbf{-.220} & \textit{\textbf{-7.6$\boldsymbol{E}$-03}} \\
Icelandic & .233 & -2.5$\mathrm{E}$-04 & .054 & .134 \\
Italian & \textbf{.147} & \textbf{-5.3$\mathbf{E}$-03} & \textbf{-.307} & 4.3$\mathrm{E}$-03 \\
Japanese & .272 & -.072 & .145 & \textcolor{red}{\textbf{.316}} \\
Latvian & \textcolor{red}{\textbf{.468}} & \textcolor{red}{\textbf{.125}} & \textbf{-.148} & \textcolor{red}{\textbf{.287}} \\
Lithuanian & .348 & .252 & -.190 & \textit{.230} \\
Norwegian & .383 & .332 & \textbf{.045} & .330 \\
Polish & .168 & \textbf{-.130} & \textbf{-.217} & \textbf{-.077} \\
Portuguese & \textbf{.088} & .041 & \textbf{.032} & .141 \\
Romanian & \textbf{.041} & \textbf{-.102} & \textbf{-.041} & \textbf{.082} \\
Russian & \textbf{.058} & \textbf{6.9$\mathbf{E}$-03} & \textbf{-.104} & \textbf{.046} \\
Slovak & .484 & .371 & \textbf{-.129} & \textit{.235} \\
Slovenian & \textbf{9.5$\mathbf{E}$-03} & \textbf{-.129} & \textbf{-.082} & .079 \\
Spanish & \textbf{4.4$\mathbf{E}$-04} & \textbf{-.085} & \textbf{-.323} & \textit{-.245} \\
Swedish & \textbf{-.120} & \textbf{-.170} & \textbf{-.299} & \textbf{-.037} \\
Turkish & -.089 & \textbf{-.190} & \textbf{-.208} & .010 \\
Ukrainian & \textbf{-.092} & \textbf{-.174} & \textbf{-.092} & 2.5$\mathrm{E}$-03 \\
\bottomrule
\end{tabularx}
\caption{Per-language estimates $\hat D = \hat\mu_A - \hat\mu_B$ on similarity-matrix correlations after UMAP projection of the embedding space onto $\mathbb{R}^{16}$ for the four invariance hypotheses at $\kappa = 1$. \textbf{Bold black}: invariance / equivalence. \textcolor{red}{\textbf{Red bold}}: the translation pipeline exhibits greater distortion than the reference pipeline. \textcolor{blue}{\textbf{Blue bold}}: the translation pipeline exhibits lesser distortion than the reference pipeline (OM--OT only). Plain: indeterminate. \textit{Italicized} OM-OT entries indicate languages for which the mixed-model fit failed to converge and the reported estimate is from the cluster-bootstrap backend instead.}
\label{tbl:allHypothesesUmap}
\end{table}

\begin{table}[t]
\centering
\small
\setlength{\tabcolsep}{4pt}
\begin{tabularx}{\columnwidth}{@{}Xlrrrr@{}}
\toprule
hypothesis & $\kappa$ & invariant & indeterm. & distort. & super. \\
\midrule
Baseline OO-OT & 0.5 & 6 & 19 & 3 & \\
 & \textbf{1.0} & \textbf{13} & \textbf{14} & \textbf{1} & \\
 & 1.5 & 16 & 11 & 1 & \\
 & 2.0 & 16 & 11 & 1 & \\
\midrule
Best model OO-OT$_\text{best}$ & 0.5 & 13 & 12 & 3 & \\
 & \textbf{1.0} & \textbf{15} & \textbf{11} & \textbf{2} & \\
 & 1.5 & 19 & 8 & 1 & \\
 & 2.0 & 24 & 3 & 1 & \\
\midrule
Multilingual MM-MX & 0.5 & 19 & 9 & 0 & \\
 & \textbf{1.0} & \textbf{24} & \textbf{4} & \textbf{0} & \\
 & 1.5 & 26 & 2 & 0 & \\
 & 2.0 & 27 & 1 & 0 & \\
\midrule
OM-OT perf. equiv. & 0.5 & 4 & 20 & 4 & 0 \\
 & \textbf{1.0} & \textbf{11} & \textbf{15} & \textbf{2} & \textbf{0} \\
 & 1.5 & 14 & 13 & 1 & 0 \\
 & 2.0 & 17 & 10 & 1 & 0 \\
\bottomrule
\end{tabularx}
\caption{Decision counts for the UMAP dimensionality-reduction analysis across all 28 languages for each hypothesis at four invariance-margin tolerances $\kappa$. The primary specification ($\kappa = 1$) is highlighted in bold. For the three directional hypotheses (top three blocks), decisions are \emph{invariant} / \emph{indeterminate} / \emph{distorted}. For the two-sided OM-OT hypothesis (bottom block), decisions are \emph{equivalent} / \emph{indeterminate} / \emph{OT inferior} / \emph{OT superior}; verdicts come from the mixed-model fit for 23 languages and from cluster-bootstrap inference for the five (Spanish, Estonian, Hungarian, Lithuanian, Slovak) for which the mixed-model fit failed to converge.}
\label{tbl:sensitivityUmap}
\end{table}

\section{Conclusions}

We introduce a new methodological framework for evaluating cross-lingual translation pipelines, with per-language verdicts that distinguish languages where translation demonstrably preserves semantic structure, languages where it demonstrably degrades it, and languages where the available evidence does not resolve the question. In particular, rather than measuring translation-induced semantic shift directly -- which would require ground-truth semantic similarity that no model perfectly captures -- we measure the \emph{stability of pairwise similarity relationships} across embedding models, by computing correlations between similarity matrices derived from original-language and translated texts. This shifts the inferential target from absolute semantic preservation (which is not practically measurable) to relative structural preservation (which is). Second, we use \emph{inter-model disagreement on original-language text} as the standard against which translation-induced distortion is judged.

Empirically, our analysis of the political manifesto corpus translated through the EU eTranslation service to English yields a clear stratification of the 28 languages we examined. Five languages -- German, Italian, Portuguese, Spanish, and Ukrainian -- exhibit full translation invariance: the translation pipeline is statistically indistinguishable from both the original-language and the multilingual baselines. Five further languages -- Bulgarian, Greek, Hungarian, Icelandic, and Swedish -- exhibit invariance on most statistical tests, but single ones fail for lack of statistical power. On the other hand, for four languages -- French, Japanese, Latvian, and Lithuanian -- the translation pipeline distorts the similarity structure relative to the original-language baseline even when the best post-translation embedding model is selected. The remaining thirteen languages are predominantly indeterminate, reflecting limited statistical power rather than positive evidence of either invariance or distortion. We cannot definitively explain these interlingual differences with our current dataset; whether they reflect MT model coverage, target-side training data scarcity, or language-typological factors is a question that requires comparison across MT systems and target languages, which our framework can support but our current data cannot answer.

On the other hand, multilingual encoders applied directly to translated text are robust across all languages in our study. When comparing original-language and translated similarity matrices obtained under the same multilingual models, translation invariance holds for $24$ of $28$ languages, and no language exhibits statistically significant difference. This suggests that a meaningful share of the apparent translation effect in the monolingual pipeline comparison is attributable to asymmetry between heterogeneous original-language models and homogeneous post-translation ones, rather than to translation per se.

The downstream-task analyses provide a more nuanced picture and qualify the generality of our findings. Translation-induced perturbations to embedding geometry do not propagate uniformly to downstream behavior. In classification tasks, systematic outcome distortions have not been detected for any language; in clustering tasks, such distortions appear only for French and Japanese; and in dimensionality reduction tasks, they also arise for Latvian and Czech. The progressive dissolution of the translation distortion cluster as the analysis moves from raw embedding geometry to coarse downstream decisions indicates that applications relying on discrete classifier outputs may be substantially less sensitive to translation distortion than applications relying on the relative geometry of representations.

Future work will extend the analysis along three axes for which our current dataset cannot provide evidence: comparison across MT systems to separate translation-quality effects from translation-process effects; comparison across pivot languages, including non-English pivots and direct cross-lingual comparisons that bypass pivot translation entirely; and extension to other corpus domains where the pattern of source-language heterogeneity differs (literary text, scientific abstracts, conversational data). Each of these extensions can be conducted within the same methodological framework, and the resulting comparisons should help identify which of our specific empirical findings -- particularly the four-language distorted cluster and the languages with positive invariance verdicts -- generalize and which are artifacts of the eTranslation-on-political-manifestos configuration we examined.

\begin{ack}
    This research has been funded under the Polish National Center for Science grant no.~2023/49/B/HS5/03893 and the Jagiellonian University Excellence Initiative, DigiWorld PRA, QuantPol Center project.
\end{ack}

\bibliography{partysim}

\pagebreak
\onecolumn
\section*{Appendix 1: Models used for generating sentence embeddings}

\begin{longtable}{lp{0.7\textwidth}}
\caption{Sentence-embedding models used per language, plus the post-translation English and multilingual model sets.}
\label{tbl:models}\\
\toprule
\textbf{Language} & \textbf{Model} \\
\midrule
\endfirsthead

\multicolumn{2}{c}{\tablename\ \thetable\ -- \textit{continued from previous page}} \\
\toprule
\textbf{Language} & \textbf{Model} \\
\midrule
\endhead

\midrule
\multicolumn{2}{r}{\textit{continued on next page}} \\
\endfoot

\bottomrule
\endlastfoot

\multirow{6}{*}{Bulgarian} & \texttt{DeepPavlov/bert-base-bg-cs-pl-ru-cased} \\
 & \texttt{iarfmoose/roberta-base-bulgarian} \\
 & \texttt{mboyanov/bg2vec} \\
 & \texttt{rmihaylov/bert-base-bg} \\
 & \texttt{rmihaylov/roberta-base-nli-stsb-bg} \\
 & \texttt{rmihaylov/roberta-base-nli-stsb-theseus-bg} \\
\midrule
\multirow{3}{*}{Croatian} & \texttt{EMBEDDIA/crosloengual-bert} \\
 & \texttt{InfoCoV/Cro-CoV-cseBERT} \\
 & \texttt{classla/bcms-bertic} \\
\midrule
\multirow{5}{*}{Czech} & \texttt{Seznam/simcse-dist-mpnet-czeng-cs-en} \\
 & \texttt{Seznam/simcse-dist-mpnet-paracrawl-cs-en} \\
 & \texttt{UWB-AIR/Czert-B-base-cased} \\
 & \texttt{fav-kky/FERNET-C5} \\
 & \texttt{ufal/robeczech-base} \\
\midrule
\multirow{7}{*}{Danish} & \texttt{DDSC/roberta-base-danish} \\
 & \texttt{KennethEnevoldsen/dfm-sentence-encoder-large-exp1} \\
 & \texttt{KennethTM/MiniLM-L6-danish-encoder-v2} \\
 & \texttt{Maltehb/danish-bert-botxo} \\
 & \texttt{alexanderfalk/danbert-small-cased} \\
 & \texttt{vesteinn/DanskBERT} \\
 & \texttt{vesteinn/ScandiBERT} \\
\midrule
\multirow{5}{*}{Dutch} & \texttt{DTAI-KULeuven/robbert-2023-dutch-large} \\
 & \texttt{NetherlandsForensicInstitute/robbert-2022-dutch-sentence-transformers} \\
 & \texttt{Parallia/Fairly-Multilingual-ModernBERT-Embed-BE-NL} \\
 & \texttt{T-Systems-onsite/cross-en-nl-roberta-sentence-transformer} \\
 & \texttt{jegormeister/bert-base-dutch-cased-snli} \\
\midrule
\multirow{4}{*}{Estonian} & \texttt{EMBEDDIA/est-roberta} \\
 & \texttt{EMBEDDIA/finest-bert} \\
 & \texttt{HPLT/hplt\_bert\_base\_2\_0\_est-Latn} \\
 & \texttt{kiri-ai/distiluse-base-multilingual-cased-et} \\
\midrule
\multirow{5}{*}{Finnish} & \texttt{EMBEDDIA/finest-bert} \\
 & \texttt{Finnish-NLP/Ahma-7B} \\
 & \texttt{Finnish-NLP/roberta-large-finnish} \\
 & \texttt{Finnish-NLP/t5-base-nl36-finnish} \\
 & \texttt{TurkuNLP/sbert-cased-finnish-paraphrase} \\
\midrule
\multirow{5}{*}{French} & \texttt{Lajavaness/bilingual-embedding-large} \\
 & \texttt{Lajavaness/sentence-flaubert-base} \\
 & \texttt{dangvantuan/sentence-camembert-large} \\
 & \texttt{manu/bge-m3-custom-fr} \\
 & \texttt{manu/sentence\_croissant\_alpha\_v0.4} \\
\midrule
\multirow{8}{*}{German} & \texttt{LennartKeller/longformer-gottbert-base-8192-aw512} \\
 & \texttt{T-Systems-onsite/cross-en-de-roberta-sentence-transformer} \\
 & \texttt{aari1995/German\_Semantic\_STS\_V2} \\
 & \texttt{aari1995/German\_Semantic\_V3b} \\
 & \texttt{deepset/gbert-large} \\
 & \texttt{google-bert/bert-base-german-cased} \\
 & \texttt{jinaai/jina-embeddings-v2-base-de} \\
 & \texttt{mixedbread-ai/deepset-mxbai-embed-de-large-v1} \\
 \pagebreak
 \multirow{5}{*}{Greek} & \texttt{DGurgurov/mbert\_ell-grek} \\
 & \texttt{DGurgurov/xlm-r\_ell-grek} \\
 & \texttt{HPLT/hplt\_bert\_base\_2\_0\_ell-Grek} \\
 & \texttt{dimitriz/st-greek-media-bert-base-uncased} \\
 & \texttt{lighteternal/stsb-xlm-r-greek-transfer} \\
\midrule
\multirow{4}{*}{Hungarian} & \texttt{NYTK/sentence-transformers-experimental-hubert-hungarian} \\
 & \texttt{SZTAKI-HLT/hubert-base-cc} \\
 & \texttt{SZTAKI-HLT/mT5-large-HuAMR} \\
 & \texttt{karsar/ModernBERT-base-hu\_v3} \\
\midrule
\multirow{5}{*}{Icelandic} & \texttt{Sigurdur/isl-sbert-l} \\
 & \texttt{jonfd/convbert-base-igc-is} \\
 & \texttt{jonfd/electra-base-igc-is} \\
 & \texttt{neurocode/IsRoBERTa} \\
 & \texttt{vesteinn/ScandiBERT} \\
\midrule
\multirow{4}{*}{Italian} & \texttt{DeepMount00/Italian-ModernBERT-base} \\
 & \texttt{efederici/sentence-BERTino} \\
 & \texttt{nickprock/multi-sentence-BERTino} \\
 & \texttt{nickprock/sentence-bert-base-italian-xxl-uncased} \\
\midrule
\multirow{6}{*}{Japanese} & \texttt{MU-Kindai/SBERT-JSNLI-large} \\
 & \texttt{cl-nagoya/ruri-large} \\
 & \texttt{cl-nagoya/sup-simcse-ja-base} \\
 & \texttt{colorfulscoop/sbert-base-ja} \\
 & \texttt{pkshatech/GLuCoSE-base-ja-v2} \\
 & \texttt{pkshatech/RoSEtta-base-ja} \\
\midrule
\multirow{3}{*}{Latvian} & \texttt{AiLab-IMCS-UL/lvbert} \\
 & \texttt{EMBEDDIA/litlat-bert} \\
 & \texttt{HPLT/hplt\_bert\_base\_2\_0\_lvs-Latn} \\
\midrule
\multirow{4}{*}{Lithuanian} & \texttt{EMBEDDIA/litlat-bert} \\
 & \texttt{Geotrend/bert-base-lt-cased} \\
 & \texttt{HPLT/hplt\_bert\_base\_2\_0\_lit-Latn} \\
 & \texttt{jkeruotis/LitBERTa-uncased} \\
\midrule
\multirow{4}{*}{Norwegian} & \texttt{FFI/SimCSE-NB-BERT-large} \\
 & \texttt{NbAiLab/nb-bert-large} \\
 & \texttt{ltgoslo/norbert2} \\
 & \texttt{vesteinn/ScandiBERT} \\
\midrule
\multirow{5}{*}{Polish} & \texttt{Voicelab/sbert-large-cased-pl} \\
 & \texttt{ipipan/silver-retriever-base-v1} \\
 & \texttt{sdadas/mmlw-roberta-large} \\
 & \texttt{sdadas/st-polish-paraphrase-from-mpnet} \\
 & \texttt{sdadas/stella-pl} \\
\midrule
\multirow{4}{*}{Portuguese} & \texttt{PORTULAN/albertina-100m-portuguese-ptpt-encoder} \\
 & \texttt{PORTULAN/serafim-335m-portuguese-pt-sentence-encoder} \\
 & \texttt{neuralmind/bert-base-portuguese-cased} \\
 & \texttt{rufimelo/bert-large-portuguese-cased-sts} \\
\midrule
\multirow{5}{*}{Romanian} & \texttt{BlackKakapo/cupidon-base-ro} \\
 & \texttt{BlackKakapo/stsb-xlm-r-multilingual-ro} \\
 & \texttt{dumitrescustefan/mt5-large-romanian} \\
 & \texttt{iliemihai/romanian-sentence-bert-base-uncased-v1} \\
 & \texttt{iliemihai/romanian-sentence-e5-large} \\
\pagebreak
\multirow{6}{*}{Russian} & \texttt{DeepPavlov/rubert-base-cased-sentence} \\
 & \texttt{T-Systems-onsite/cross-en-ru-roberta-sentence-transformer} \\
 & \texttt{ai-forever/ru-en-RoSBERTa} \\
 & \texttt{ai-forever/sbert\_large\_mt\_nlu\_ru} \\
 & \texttt{ai-forever/sbert\_large\_nlu\_ru} \\
 & \texttt{sergeyzh/LaBSE-ru-sts} \\
\midrule
\multirow{4}{*}{Slovak} & \texttt{HPLT/hplt\_bert\_base\_2\_0\_slk-Latn} \\
 & \texttt{fav-kky/FERNET-News\_sk} \\
 & \texttt{gerulata/slovakbert} \\
 & \texttt{kinit/slovakbert-sts-stsb} \\
\midrule
\multirow{5}{*}{Slovenian} & \texttt{EMBEDDIA/crosloengual-bert} \\
 & \texttt{EMBEDDIA/sloberta} \\
 & \texttt{cjvt/sloberta-sleng} \\
 & \texttt{cjvt/t5-sl-large} \\
 & \texttt{rokn/slovlo-v1} \\
\midrule
\multirow{6}{*}{Spanish} & \texttt{DeepESP/gpt2-spanish} \\
 & \texttt{PlanTL-GOB-ES/longformer-base-4096-bne-es} \\
 & \texttt{PlanTL-GOB-ES/roberta-large-bne} \\
 & \texttt{T-Systems-onsite/cross-en-es-roberta-sentence-transformer} \\
 & \texttt{jinaai/jina-embeddings-v2-base-es} \\
 & \texttt{skimai/spanberta-base-cased} \\
\midrule
\multirow{5}{*}{Swedish} & \texttt{Contrastive-Tension/BERT-Base-Swe-CT-STSb} \\
 & \texttt{KBLab/sentence-bert-swedish-cased} \\
 & \texttt{jekunz/modernbert-fineweb-sv} \\
 & \texttt{jzju/sbert-sv-lim2} \\
 & \texttt{vesteinn/ScandiBERT} \\
\midrule
\multirow{5}{*}{Turkish} & \texttt{atasoglu/turkish-base-bert-uncased-mean-nli-stsb-tr} \\
 & \texttt{dbmdz/bert-base-turkish-cased} \\
 & \texttt{emrecan/convbert-base-turkish-mc4-cased-allnli\_tr} \\
 & \texttt{eneSadi/turkuaz-embeddings} \\
 & \texttt{myzens/turemb\_512} \\
\midrule
\multirow{5}{*}{Ukrainian} & \texttt{HPLT/hplt\_bert\_base\_2\_0\_ukr-Cyrl} \\
 & \texttt{ai-forever/mGPT-1.3B-ukranian} \\
 & \texttt{lang-uk/ukr-paraphrase-multilingual-mpnet-base} \\
 & \texttt{maiia-bocharova/ukr\_sentence\_mpnet\_cos\_sim} \\
 & \texttt{panalexeu/xlm-roberta-ua-distilled} \\
\midrule
\multirow{6}{*}{translated} & \texttt{sentence-transformers/stsb-bert-large} \\
 & \texttt{sentence-transformers/nli-mpnet-base-v2} \\
 & \texttt{WhereIsAI/UAE-Large-V1} \\
 & \texttt{intfloat/e5-mistral-7b-instruct} \\
 & \texttt{NousResearch/Llama-2-7b-hf} \\
 & \texttt{nvidia/NV-Embed-v2} \\
\midrule
\multirow{7}{*}{multilingual} & \texttt{sentence-transformers/paraphrase-multilingual-mpnet-base-v2} \\
 & \texttt{FacebookAI/xlm-roberta-large} \\
 & \texttt{markussagen/xlm-roberta-longformer-base-4096} \\
 & \texttt{sentence-transformers/LaBSE} \\
 & \texttt{google/mt5-large} \\
 & \texttt{nomic-ai/nomic-embed-text-v1.5} \\
 & \texttt{BAAI/bge-m3} \\
\end{longtable}

\twocolumn
\section*{Appendix 2: Detailed results}

\begin{table}[hb]
\centering
\small
\setlength{\tabcolsep}{3pt}
\begin{tabularx}{\columnwidth}{@{}Xrrrrcrl@{}}
\toprule
language & $n$ & $n_O$ & $n_T$ & $\hat D$ & $90\%$ CI & $p$-value & dec. \\
\midrule
Bulgarian & 114 & 6 & 7 & \phantom{-}.14 & (\phantom{-}.02, \phantom{-}.25) & $1.8\mathrm{E}-29$ & inv. \\
Croatian & 68 & 4 & 7 & \phantom{-}.27 & (\phantom{-}.16, \phantom{-}.38) & $9.5\mathrm{E}-01$ & ind. \\
Czech & 90 & 5 & 7 & \phantom{-}.25 & (\phantom{-}.14, \phantom{-}.35) & $9.2\mathrm{E}-01$ & ind. \\
Danish & 140 & 7 & 7 & \phantom{-}.06 & (-.34, \phantom{-}.46) & $2.9\mathrm{E}-01$ & ind. \\
Dutch & 80 & 5 & 6 & \phantom{-}.14 & (\phantom{-}.09, \phantom{-}.19) & $6.7\mathrm{E}-01$ & ind. \\
English & 36 & 3 & 5 & -.05 & (-.30, \phantom{-}.19) & $2.6\mathrm{E}-01$ & ind. \\
Estonian & 68 & 4 & 7 & \phantom{-}.19 & (-.37, \phantom{-}.75) & $4.7\mathrm{E}-01$ & ind. \\
Finnish & 90 & 5 & 7 & \phantom{-}.06 & (-.31, \phantom{-}.43) & $2.6\mathrm{E}-01$ & ind. \\
French & 90 & 5 & 7 & \phantom{-}.20 & (\phantom{-}.07, \phantom{-}.34) & $9.6\mathrm{E}-01$ & dis. \\
German & 168 & 8 & 7 & \phantom{-}.03 & (-.34, \phantom{-}.39) & $9.8\mathrm{E}-05$ & inv. \\
Greek & 90 & 5 & 7 & \phantom{-}.07 & (-.09, \phantom{-}.23) & $6.7\mathrm{E}-04$ & inv. \\
Hungarian & 68 & 4 & 7 & \phantom{-}.10 & (-.21, \phantom{-}.41) & $1.0\mathrm{E}-01$ & ind. \\
Icelandic & 90 & 5 & 7 & \phantom{-}.15 & (-.14, \phantom{-}.44) & $1.2\mathrm{E}-01$ & ind. \\
Italian & 68 & 4 & 7 & \phantom{-}.00 & (-.09, \phantom{-}.10) & $4.0\mathrm{E}-07$ & inv. \\
Japanese & 114 & 6 & 7 & \phantom{-}.42 & (\phantom{-}.30, \phantom{-}.55) & $1.0\mathrm{E}+00$ & dis. \\
Latvian & 48 & 3 & 7 & \phantom{-}.23 & (\phantom{-}.13, \phantom{-}.33) & $1.0\mathrm{E}+00$ & dis. \\
Lithuanian & 68 & 4 & 7 & \phantom{-}.35 & (\phantom{-}.25, \phantom{-}.45) & $1.0\mathrm{E}+00$ & dis. \\
Norwegian & 68 & 4 & 7 & \phantom{-}.30 & (\phantom{-}.20, \phantom{-}.41) & $8.4\mathrm{E}-01$ & ind. \\
Polish & 90 & 5 & 7 & \phantom{-}.14 & (-.01, \phantom{-}.29) & $8.4\mathrm{E}-01$ & ind. \\
Portuguese & 68 & 4 & 7 & \phantom{-}.17 & (-.10, \phantom{-}.43) & $9.2\mathrm{E}-03$ & inv. \\
Romanian & 90 & 5 & 7 & \phantom{-}.09 & (-.43, \phantom{-}.61) & $3.6\mathrm{E}-01$ & ind. \\
Russian & 114 & 6 & 7 & \phantom{-}.06 & (-.01, \phantom{-}.13) & $2.8\mathrm{E}-01$ & ind. \\
Slovak & 68 & 4 & 7 & \phantom{-}.30 & (-.29, \phantom{-}.90) & $5.7\mathrm{E}-01$ & ind. \\
Slovenian & 90 & 5 & 7 & \phantom{-}.26 & (-.12, \phantom{-}.64) & $6.7\mathrm{E}-01$ & ind. \\
Spanish & 114 & 6 & 7 & \phantom{-}.16 & (-.46, \phantom{-}.78) & $1.0\mathrm{E}-02$ & inv. \\
Swedish & 90 & 5 & 7 & -.09 & (-.70, \phantom{-}.53) & $2.3\mathrm{E}-04$ & inv. \\
Turkish & 90 & 5 & 7 & \phantom{-}.01 & (-.47, \phantom{-}.49) & $2.5\mathrm{E}-01$ & ind. \\
Ukrainian & 90 & 5 & 7 & \phantom{-}.03 & (-.33, \phantom{-}.40) & $4.6\mathrm{E}-03$ & inv. \\
\bottomrule
\end{tabularx}
\caption{Per-language test outcomes for the baseline invariance hypothesis (OO vs OT) at $\kappa = 1$. Decision codes: inv.\ (invariant), dis.\ (distorted), ind.\ (indeterminate). $\hat D = \hat\mu_{OO} - \hat\mu_{OT}$; $p$ is the one-sided non-inferiority $p$-value.}
\label{tbl:baselineInvariance}
\end{table}

\begin{table}[hb]
\centering
\small
\setlength{\tabcolsep}{3pt}
\begin{tabularx}{\columnwidth}{@{}Xrrrrcrl@{}}
\toprule
language & $n$ & $n_O$ & $n_T$ & $\hat D$ & $90\%$ CI & $p$-value & dec. \\
\midrule
Bulgarian & 114 & 6 & 7 & \phantom{-}.06 & (-.04, \phantom{-}.22) & $<\!10^{-30}$ & inv. \\
Croatian & 68 & 4 & 7 & \phantom{-}.18 & (\phantom{-}.09, \phantom{-}.30) & $6.6\mathrm{E}-01$ & ind. \\
Czech & 90 & 5 & 7 & \phantom{-}.18 & (\phantom{-}.09, \phantom{-}.26) & $7.2\mathrm{E}-01$ & ind. \\
Danish & 140 & 7 & 7 & \phantom{-}.01 & (-.11, \phantom{-}.16) & $2.3\mathrm{E}-02$ & inv. \\
Dutch & 80 & 5 & 6 & \phantom{-}.12 & (\phantom{-}.06, \phantom{-}.18) & $3.7\mathrm{E}-01$ & ind. \\
English & 36 & 3 & 5 & -.08 & (-.14, -.03) & $2.8\mathrm{E}-03$ & inv. \\
Estonian & 68 & 4 & 7 & \phantom{-}.15 & (-.11, \phantom{-}.38) & $3.0\mathrm{E}-01$ & ind. \\
Finnish & 90 & 5 & 7 & \phantom{-}.03 & (-.13, \phantom{-}.17) & $9.1\mathrm{E}-03$ & inv. \\
French & 90 & 5 & 7 & \phantom{-}.18 & (\phantom{-}.12, \phantom{-}.23) & $1.0\mathrm{E}+00$ & dis. \\
German & 168 & 8 & 7 & \phantom{-}.00 & (-.05, \phantom{-}.07) & $<\!10^{-30}$ & inv. \\
Greek & 90 & 5 & 7 & -.07 & (-.22, \phantom{-}.12) & $5.0\mathrm{E}-04$ & inv. \\
Hungarian & 68 & 4 & 7 & \phantom{-}.05 & (-.03, \phantom{-}.15) & $<\!10^{-30}$ & inv. \\
Icelandic & 90 & 5 & 7 & \phantom{-}.06 & (-.03, \phantom{-}.22) & $3.9\mathrm{E}-03$ & inv. \\
Italian & 68 & 4 & 7 & -.03 & (-.20, \phantom{-}.11) & $<\!10^{-30}$ & inv. \\
Japanese & 114 & 6 & 7 & \phantom{-}.17 & (\phantom{-}.15, \phantom{-}.39) & $1.0\mathrm{E}+00$ & dis. \\
Latvian & 48 & 3 & 7 & \phantom{-}.13 & (\phantom{-}.07, \phantom{-}.20) & $1.0\mathrm{E}+00$ & dis. \\
Lithuanian & 68 & 4 & 7 & \phantom{-}.28 & (\phantom{-}.25, \phantom{-}.32) & $1.0\mathrm{E}+00$ & dis. \\
Norwegian & 68 & 4 & 7 & \phantom{-}.28 & (-.04, \phantom{-}.56) & $5.7\mathrm{E}-01$ & ind. \\
Polish & 90 & 5 & 7 & \phantom{-}.02 & (-.01, \phantom{-}.07) & $1.9\mathrm{E}-01$ & ind. \\
Portuguese & 68 & 4 & 7 & \phantom{-}.13 & (-.01, \phantom{-}.30) & $<\!10^{-30}$ & inv. \\
Romanian & 90 & 5 & 7 & -.02 & (-.14, \phantom{-}.18) & $4.4\mathrm{E}-02$ & inv. \\
Russian & 114 & 6 & 7 & \phantom{-}.00 & (-.06, \phantom{-}.07) & $3.5\mathrm{E}-02$ & inv. \\
Slovak & 68 & 4 & 7 & \phantom{-}.12 & (-.07, \phantom{-}.52) & $2.2\mathrm{E}-01$ & ind. \\
Slovenian & 90 & 5 & 7 & \phantom{-}.18 & (\phantom{-}.03, \phantom{-}.36) & $6.2\mathrm{E}-01$ & ind. \\
Spanish & 114 & 6 & 7 & \phantom{-}.11 & (-.05, \phantom{-}.43) & $<\!10^{-30}$ & inv. \\
Swedish & 90 & 5 & 7 & -.14 & (-.28, \phantom{-}.00) & $<\!10^{-30}$ & inv. \\
Turkish & 90 & 5 & 7 & -.05 & (-.17, \phantom{-}.14) & $1.0\mathrm{E}-02$ & inv. \\
Ukrainian & 90 & 5 & 7 & -.02 & (-.12, \phantom{-}.05) & $<\!10^{-30}$ & inv. \\
\bottomrule
\end{tabularx}
\caption{Per-language test outcomes for the best-model invariance hypothesis (OO vs OT$_\text{best}$) at $\kappa = 1$. Decision codes: inv.\ (invariant), dis.\ (distorted), ind.\ (indeterminate). $\hat D = \hat\mu_{OO} - \hat\mu_{OT_\text{best}}$; $p$ is the one-sided non-inferiority $p$-value. Confidence intervals are computed via configuration-level cluster bootstrap that re-selects the best post-translation model within each replicate.}
\label{tbl:bestmodelInvariance}
\end{table}

\begin{table}[t]
\centering
\small
\setlength{\tabcolsep}{3pt}
\begin{tabularx}{\columnwidth}{@{}Xrrrrcrl@{}}
\toprule
language & $n$ & $n_O$ & $n_T$ & $\hat D$ & $90\%$ CI & $p$-value & dec. \\
\midrule
Bulgarian & 56 & 6 & 7 & -.05 & (-.52, \phantom{-}.43) & $2.0\mathrm{E}-01$ & ind. \\
Croatian & 56 & 4 & 7 & -.10 & (-.45, \phantom{-}.24) & $2.1\mathrm{E}-04$ & inv. \\
Czech & 56 & 5 & 7 & -.07 & (-.19, \phantom{-}.04) & $3.6\mathrm{E}-84$ & inv. \\
Danish & 56 & 7 & 7 & -.07 & (-.54, \phantom{-}.39) & $4.8\mathrm{E}-04$ & inv. \\
Dutch & 56 & 5 & 6 & -.03 & (-.55, \phantom{-}.49) & $8.2\mathrm{E}-04$ & inv. \\
English & 46 & 3 & 5 & -.37 & (-1.55, \phantom{-}.82) & $9.8\mathrm{E}-03$ & inv. \\
Estonian & 56 & 4 & 7 & -.16 & (-.68, \phantom{-}.35) & $1.8\mathrm{E}-04$ & inv. \\
Finnish & 56 & 5 & 7 & -.24 & (-.67, \phantom{-}.19) & $5.5\mathrm{E}-02$ & ind. \\
French & 56 & 5 & 7 & -.05 & (-.62, \phantom{-}.52) & $6.9\mathrm{E}-04$ & inv. \\
German & 56 & 8 & 7 & \phantom{-}.03 & (-.10, \phantom{-}.17) & $2.4\mathrm{E}-32$ & inv. \\
Greek & 56 & 5 & 7 & \phantom{-}.09 & (-.13, \phantom{-}.30) & $8.1\mathrm{E}-02$ & ind. \\
Hungarian & 56 & 4 & 7 & -.05 & (-.45, \phantom{-}.36) & $6.5\mathrm{E}-04$ & inv. \\
Icelandic & 56 & 5 & 7 & -.11 & (-.49, \phantom{-}.26) & $2.0\mathrm{E}-04$ & inv. \\
Italian & 56 & 4 & 7 & -.18 & (-.77, \phantom{-}.42) & $2.0\mathrm{E}-04$ & inv. \\
Japanese & 56 & 6 & 7 & -.20 & (-.84, \phantom{-}.44) & $1.8\mathrm{E}-04$ & inv. \\
Latvian & 56 & 3 & 7 & -.18 & (-.56, \phantom{-}.20) & $7.0\mathrm{E}-05$ & inv. \\
Lithuanian & 56 & 4 & 7 & -.10 & (-.78, \phantom{-}.58) & $5.1\mathrm{E}-04$ & inv. \\
Norwegian & 56 & 4 & 7 & -.03 & (-.68, \phantom{-}.62) & $8.9\mathrm{E}-04$ & inv. \\
Polish & 56 & 5 & 7 & -.22 & (-.30, -.14) & $4.6\mathrm{E}-16$ & inv. \\
Portuguese & 56 & 4 & 7 & \phantom{-}.01 & (-.50, \phantom{-}.52) & $1.4\mathrm{E}-03$ & inv. \\
Romanian & 56 & 5 & 7 & -.13 & (-.21, -.05) & $4.7\mathrm{E}-64$ & inv. \\
Russian & 56 & 6 & 7 & -.09 & (-.53, \phantom{-}.34) & $3.8\mathrm{E}-04$ & inv. \\
Slovak & 56 & 4 & 7 & -.01 & (-.52, \phantom{-}.51) & $1.1\mathrm{E}-03$ & inv. \\
Slovenian & 56 & 5 & 7 & -.10 & (-.78, \phantom{-}.58) & $5.2\mathrm{E}-04$ & inv. \\
Spanish & 56 & 6 & 7 & -.11 & (-.71, \phantom{-}.49) & $4.1\mathrm{E}-04$ & inv. \\
Swedish & 56 & 5 & 7 & -.25 & (-.71, \phantom{-}.21) & $6.3\mathrm{E}-02$ & ind. \\
Turkish & 56 & 5 & 7 & -.14 & (-.65, \phantom{-}.38) & $2.7\mathrm{E}-04$ & inv. \\
Ukrainian & 56 & 5 & 7 & -.09 & (-.56, \phantom{-}.38) & $4.2\mathrm{E}-04$ & inv. \\
\bottomrule
\end{tabularx}
\caption{Per-language test outcomes for the multilingual invariance hypothesis (MM vs MX) at $\kappa = 1$, with the candidate class restricted to same-model pairs (each multilingual model on original-language text vs.\ the same multilingual model on translated text). Decision codes: inv.\ (invariant), dis.\ (distorted), ind.\ (indeterminate). $\hat D = \hat\mu_{MM} - \hat\mu_{MX}$; $p$ is the one-sided non-inferiority $p$-value.}
\label{tbl:multilingualInvariance}
\end{table}

\begin{table}[t]
\centering
\small
\setlength{\tabcolsep}{3pt}
\begin{tabularx}{\columnwidth}{@{}Xrrrrcrl@{}}
\toprule
language & $n$ & $n_O$ & $n_T$ & $\hat D$ & $90\%$ CI & $p$-value & dec. \\
\midrule
Bulgarian & 168 & 6 & 7 & \phantom{-}.13 & (-.27, \phantom{-}.53) & $5.5\mathrm{E}-04$ & equ. \\
Croatian & 112 & 4 & 7 & \phantom{-}.19 & (\phantom{-}.11, \phantom{-}.28) & $7.5\mathrm{E}-01$ & ind. \\
Czech & 140 & 5 & 7 & \phantom{-}.12 & (\phantom{-}.04, \phantom{-}.19) & $2.0\mathrm{E}-01$ & ind. \\
Danish & 196 & 7 & 7 & \phantom{-}.13 & (-.36, \phantom{-}.62) & $4.1\mathrm{E}-01$ & ind. \\
Dutch & 130 & 5 & 6 & \phantom{-}.03 & (-.07, \phantom{-}.13) & $4.8\mathrm{E}-02$ & equ. \\
English & 72 & 3 & 5 & -.27 & (-.47, -.06) & $9.6\mathrm{E}-01$ & sup. \\
Estonian & 112 & 4 & 7 & \textit{\phantom{-}.16} & \textit{(\phantom{-}.02, \phantom{-}.30)} & $2.3\mathrm{E}-01$ & ind. \\
Finnish & 140 & 5 & 7 & -.02 & (-.37, \phantom{-}.34) & $1.9\mathrm{E}-01$ & ind. \\
French & 140 & 5 & 7 & -.06 & (-.20, \phantom{-}.09) & $4.9\mathrm{E}-01$ & ind. \\
German & 224 & 8 & 7 & \phantom{-}.01 & (-.47, \phantom{-}.49) & $2.1\mathrm{E}-03$ & equ. \\
Greek & 140 & 5 & 7 & \phantom{-}.15 & (-.43, \phantom{-}.73) & $2.5\mathrm{E}-01$ & ind. \\
Hungarian & 112 & 4 & 7 & \textit{\phantom{-}.09} & \textit{(-.03, \phantom{-}.21)} & $6.0\mathrm{E}-04$ & equ. \\
Icelandic & 140 & 5 & 7 & \phantom{-}.11 & (\phantom{-}.03, \phantom{-}.18) & $1.2\mathrm{E}-09$ & equ. \\
Italian & 112 & 4 & 7 & -.10 & (-.50, \phantom{-}.30) & $2.2\mathrm{E}-01$ & ind. \\
Japanese & 168 & 6 & 7 & \phantom{-}.20 & (\phantom{-}.05, \phantom{-}.35) & $9.3\mathrm{E}-01$ & ind. \\
Latvian & 84 & 3 & 7 & \phantom{-}.12 & (\phantom{-}.04, \phantom{-}.19) & $9.7\mathrm{E}-01$ & inf. \\
Lithuanian & 112 & 4 & 7 & \phantom{-}.21 & (\phantom{-}.13, \phantom{-}.29) & $1.0\mathrm{E}+00$ & inf. \\
Norwegian & 112 & 4 & 7 & \phantom{-}.21 & (\phantom{-}.13, \phantom{-}.29) & $2.6\mathrm{E}-01$ & ind. \\
Polish & 140 & 5 & 7 & -.09 & (-.24, \phantom{-}.07) & $6.6\mathrm{E}-01$ & ind. \\
Portuguese & 112 & 4 & 7 & \phantom{-}.12 & (-.29, \phantom{-}.53) & $4.4\mathrm{E}-02$ & equ. \\
Romanian & 140 & 5 & 7 & \phantom{-}.09 & (-.39, \phantom{-}.56) & $3.4\mathrm{E}-01$ & ind. \\
Russian & 168 & 6 & 7 & -.07 & (-.16, \phantom{-}.03) & $3.6\mathrm{E}-01$ & ind. \\
Slovak & 112 & 4 & 7 & \textit{\phantom{-}.32} & \textit{(\phantom{-}.17, \phantom{-}.49)} & $7.7\mathrm{E}-01$ & ind. \\
Slovenian & 140 & 5 & 7 & \phantom{-}.22 & (\phantom{-}.17, \phantom{-}.27) & $9.7\mathrm{E}-01$ & inf. \\
Spanish & 168 & 6 & 7 & \textit{\phantom{-}.14} & \textit{(\phantom{-}.02, \phantom{-}.28)} & $<\!10^{-30}$ & equ. \\
Swedish & 140 & 5 & 7 & -.01 & (-.67, \phantom{-}.64) & $1.2\mathrm{E}-03$ & equ. \\
Turkish & 140 & 5 & 7 & \phantom{-}.05 & (-.52, \phantom{-}.63) & $3.4\mathrm{E}-01$ & ind. \\
Ukrainian & 140 & 5 & 7 & \phantom{-}.02 & (-.52, \phantom{-}.56) & $3.7\mathrm{E}-02$ & equ. \\
\bottomrule
\end{tabularx}
\caption{Per-language equivalence test for the multilingual-translation performance-equivalence hypothesis (OM vs OT) at $\kappa = 1$. This hypothesis is two-sided, so decision codes are: equ.\ (equivalent), inf.\ (OT detectably inferior to OM), sup.\ (OT detectably superior to OM), ind.\ (indeterminate). $\hat D = \hat\mu_{OM} - \hat\mu_{OT}$; $p$ is the equivalence $p$-value (two one-sided tests). Italicized rows for Estonian, Hungarian, Slovak, and Spanish indicate languages for which the mixed-model fit failed to converge and the reported estimate, confidence interval, and $p$-value are from the cluster-bootstrap backend instead.}
\label{tbl:omOtEquivalence}
\end{table}

\end{document}